\documentclass[10pt]{article}

\usepackage[left=1.2in, right=1.2in, top=1.0in, bottom=1.0in, letterpaper]{geometry}
\usepackage{wrapfig}
\usepackage{color}
\usepackage[T1]{fontenc}
\usepackage[utf8]{inputenc}
\usepackage{microtype}

\usepackage{amsmath, amssymb, amsthm}

\usepackage[table]{xcolor}
\definecolor{themecolor}{RGB}{180, 100, 20}
\definecolor{rowgray}{RGB}{242, 242, 247}
\definecolor{rowbest}{RGB}{230, 242, 255}

\usepackage{graphicx}
\usepackage{subcaption}
\usepackage{booktabs}
\usepackage{multirow}
\usepackage{array}
\usepackage[labelfont={bf,color=themecolor}, labelsep=period]{caption}

\usepackage{hyperref}
\hypersetup{colorlinks=true, linkcolor=themecolor, citecolor=themecolor, urlcolor=themecolor}
\usepackage{cleveref}

\usepackage{natbib}
\setcitestyle{numbers,square,comma}

\usepackage{titlesec}
\titleformat{\section}{\large\bfseries\color{themecolor}}{\thesection}{0.7em}{}
\titleformat{\subsection}{\normalsize\bfseries\color{themecolor!80!black}}{\thesubsection}{0.6em}{}
\titleformat{\subsubsection}{\normalsize\bfseries}{\thesubsubsection}{0.6em}{}
\titlespacing*{\section}{0pt}{1.6ex plus 0.4ex}{0.6ex}
\titlespacing*{\subsection}{0pt}{1.2ex plus 0.3ex}{0.4ex}
\definecolor{cellbest}{RGB}{210, 225, 240}
\definecolor{cellsecond}{RGB}{235, 242, 248}
\makeatletter
\renewcommand{\@maketitle}{
  \vspace*{-10pt}
  {\color{themecolor}\hrule height 4pt}
  \vspace{6pt}
  \begin{center}
    {\fontsize{17}{20}\fontseries{bx}\selectfont\@title\par}
  \end{center}
  \vspace{6pt}
  {\color{themecolor}\hrule height 1pt}
  \vspace{12pt}
  \begin{center}
    {\normalsize\@author\par}
  \end{center}
  \vspace{12pt}
}
\makeatother
\usepackage[bold]{libertine}
\usepackage{abstract}

\usepackage[framemethod=tikz]{mdframed}
\mdfdefinestyle{abstractbox}{
  linecolor=themecolor,
  linewidth=1pt,
  roundcorner=16pt,
  innerleftmargin=12pt,
  innerrightmargin=12pt,
  innertopmargin=12pt,
  innerbottommargin=12pt,
  skipabove=0pt,
  skipbelow=0pt,
}
\usepackage{tikz}
\usetikzlibrary{calc}

\usepackage{enumitem}
\setlist[enumerate]{leftmargin=*, label=\arabic*.}
\setlist[itemize]{leftmargin=*}

\providecommand{\authornote}[1]{}
\providecommand{\authornotemark}[1]{}
\providecommand{\orcid}[1]{}
\providecommand{\affiliation}[1]{}
\providecommand{\institution}[1]{}
\providecommand{\city}[1]{}
\providecommand{\state}[1]{}
\providecommand{\country}[1]{}
\providecommand{\email}[1]{}
\providecommand{\keywords}[1]{}
\providecommand{\Description}[1]{}
\providecommand{\settopmatter}[1]{}
\providecommand{\acmConference}[4]{}
\providecommand{\acmYear}[1]{}
\providecommand{\acmDOI}[1]{}
\providecommand{\acmISBN}[1]{}
\providecommand{\acmSubmissionID}[1]{}
\providecommand{\copyrightyear}[1]{}
\providecommand{\setcopyright}[1]{}
\usepackage{float}
\begin{document}

% ---------- title block (article-class style) ----------
\title{\textbf{\fontsize{18}{30}\selectfont\bfseries Not All Actions Are Equal: Rethinking Conditioning for Dexterous World Model}}
\author{
  {\large\textbf{Zizhao Yuan}}$^{1}$ \;
  {\large\textbf{Zhengtu Liang}}$^{2}$ \;
  {\large\textbf{Taowen Wang}}$^{1}$ \;
  {\large\textbf{Qiwei Liang}}$^{1,2}$ \;
  {\large\textbf{Yichi Wang}}$^{3}$ \\[0.4em]
  {\large\textbf{Yunheng Wang}}$^{1}$ \;
  {\large\textbf{Yuetong Fang}}$^{1}$ \;
  {\large\textbf{Lusong Li}}$^{4}$ \;
  {\large\textbf{Zecui Zeng}}$^{4}$ \;
  {\large\textbf{Renjing Xu}}$^{1,\dagger}$ \\[1.2em]
  % {\small $^1$ HKUST(GZ) \quad $^2$ SZU \quad $^3$ BJUT \quad $^4$ JD Explore Academy} \\[0.3em]
  % {\small $^\dagger$ Corresponding author} \\[0.3em]
    {\small $^1$ The Hong Kong University of Science and Technology (Guangzhou) \\[0.3em]\quad $^2$ Shenzhen University \quad $^3$ Beijing University of Technology \quad $^4$ JD Explore Academy} \\[0.3em]
  {\small $^\dagger$ Corresponding author} \\[0.3em]
}
\date{}
\maketitle

\begin{mdframed}[style=abstractbox]
\begin{center}
{\textbf{\color{themecolor}\large Abstract}}
\end{center}
\vspace{4pt}
% Recent action-conditioned world models have begun to handle high-dimensional dexterous interactions, largely enabled by stronger visual representations and model capacity. However, the design of action conditioning itself remains largely unchanged: most methods compress actions into a single global embedding, a strategy that works well for low-dimensional control but does not generalize to high-DoF settings. We observe that dexterous actions are inherently heterogeneous, spanning multiple orders of magnitude, where large-scale motions dominate subtle but semantically critical signals. When uniformly aggregated, this leads to gradient domination, suppressing fine-grained action cues and degrading action fidelity.
% To address this, we propose DexAC, a world model that treats action conditioning as a structured process rather than global compression. We introduce a dimension-wise action tokenizer to preserve heterogeneous semantics, and a unified conditioning framework that aligns action signals with visual dynamics adaptively via local refinement and global modulation. Experiments on EgoDex show consistent improvements in action fidelity and physical consistency. Our results highlight that the key bottleneck in high-DoF world modeling lies not in model capacity, but in the mismatch between heterogeneous control signals and uniform conditioning mechanisms.
Recent advances in action-conditioned world models show promising progress in modeling complex interactions and forecasting future states under diverse action sequences. While these models are often driven by stronger visual representations and model capacity, action conditioning itself remains underexplored. Most existing approaches compress the entire action sequence into a single representation, which works well for low-DoF control but becomes less reliable in high-DoF scenarios. We observe that high-DoF dexterous actions are inherently heterogeneous, spanning multiple orders of magnitude, where large-scale motions coexist with subtle but important signals. When uniformly aggregated, optimization exhibits an imbalance across action components, which hinders the modeling of fine-grained effects and affects action fidelity. We therefore propose \textbf{DexAC-WM}, which treats action conditioning as a structured process rather than global compression. DexAC preserves dimension-level semantics via action tokenization and aligns action signals with visual dynamics through local refinement and global modulation. To address the limited high-level semantic grounding in existing world models, we further introduce a semantic branch that provides rich object-scene priors, which enables world model to capture dynamic visual details while supporting high-DoF action-conditioned video prediction. Experiments on EgoDex and EgoVerse show that combining the semantic branch with DexAC significantly improves FID, FVD, and PCK, demonstrating gains in visual-temporal realism and action-following consistency. We further verify that DexAC extends to other backbones, showing the scalability of our structured action-conditioning design. These results suggest that scaling world models to high-DoF control requires both structured action modeling and semantic grounding.

\vspace{6pt}
\noindent{\small\color{themecolor}\href{https://zizhaoyuan.github.io/DexAC-WM}{Project Page: https://zizhaoyuan.github.io/DexAC-WM}}
\end{mdframed}


\section{Introduction}
Egocentric perspective provides strong physical priors and action intent, reflecting how humans naturally see, perceive, and interact with the world~\cite{grauman2022ego4d,grauman2024ego,yu2025egosim}. To achieve human-level dexterity, it is crucial to fully leverage egocentric data so that the world model can intrinsically understand the action space from a first-person perspective. Recent progress has demonstrated strong performance in environments with relatively simple action spaces, such as navigation with discrete actions (e.g., forward, backward, and turning commands) and low-DoF robotic manipulation~\cite{bruce2024genie,wu2023daydreamer,hu2023gaia}. In these settings, models are able to generate temporally coherent and physically consistent predictions that closely follow the given actions, suggesting that action-conditioned generative modeling is a promising paradigm for learning interactive world dynamics.

\begin{figure}[t]
    \centering
    \includegraphics[width=\textwidth]{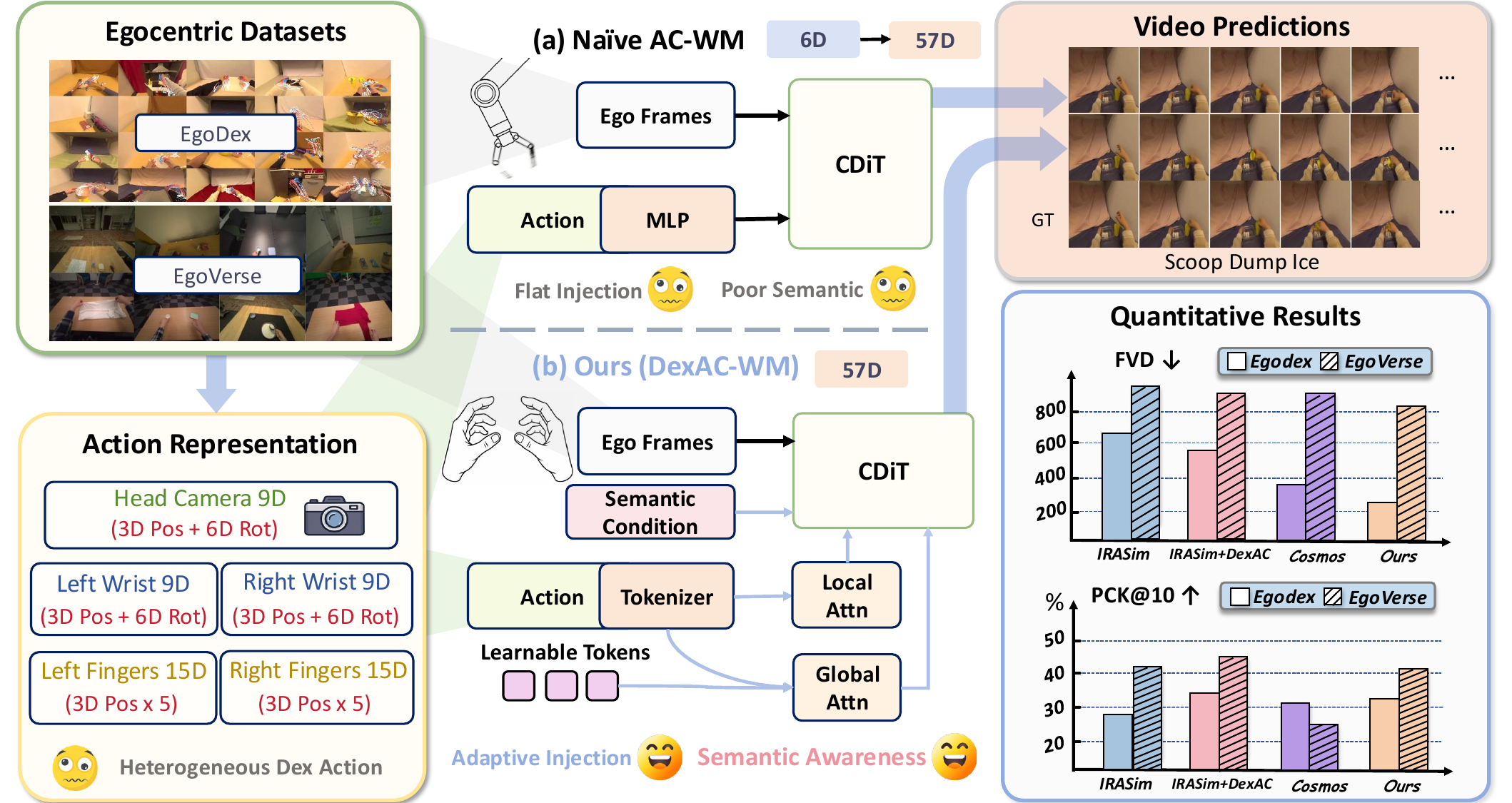}
    \caption{Overview of our DexAC-WM. Two action-conditioning paradigms for generative world models. (a) A common design in existing approaches is to aggregate the action sequence into a compact global representation before injection. While effective in low-DoF settings, such aggregation becomes less expressive in high-DoF dexterous scenarios with heterogeneous action dimensions. (b) Our proposed DexAC-WM preserves dimension-level action semantics through tokenization and combines local refinement with global modulation to better align action signals with semantic dynamics.}
    \label{fig:teaser_right}
    \vspace{-10pt}
\end{figure}

However, low-DoF success does not generalize naturally to scenarios involving high-DoF dexterous control. Unlike low-DoF world models for manipulation~\cite{hansen2023td,yang2023learning,wu2023daydreamer,guo2025ctrl}, where each action dimension contributes in a comparable magnitude across 6 DoF, dexterous human actions span multiple orders of magnitude (see Figure~\ref{fig:magnitude_comparison}), Such variability arises from the intricate structure of the human hand and its specialized motor system, which together enable motions at substantially different scales~\cite{sobinov2021neural}. In egocentric videos, global motion factors, such as wrist or camera movement, dominate in scale, yet fine-grained finger articulations remain extremely subtle yet semantically critical~\cite{goswami2026worldmodelslearningdexterous}. This imbalance introduces a gradient dominance effect, in which high-variance dimensions disproportionately influence optimization and effectively suppress low-magnitude signals. Previous research such as PEVA~\cite{bai2025whole} has preliminarily explored the inherent variability of whole-body motion, yet such investigations remain severely underexplored for dexterous world models. Consequently, how to structurally represent the subtle hand actions is particularly challenging due to: \par
\textbf{(1) Heterogeneity and subtlety in high-DoF dexterous actions.} This raises a fundamental question: how should action conditioning be structurally represented to mitigate the heterogeneous semantic collapse of dexterous hands within these complex egocentric world models? \textbf{(2) Limitations of global compression in action conditioning.} Existing studies~\cite{zheng2025survey,an2025dexterous,li2022survey} have mainly explored three forms of representation of dexterous actions. One line of work uses joint-space representations, in which re-targeted high-DoF joint values are directly taken as action inputs~\cite{wang2024dexcap,qin2022dexmv,qin2023anyteleop}. Another line of research adopts geometry-aware keypoint or fingertip representations, which are generally easier to extract from human videos~\cite{qin2023dexpoint,handa2023dextreme,wen2023any}. A third direction explores mesh-driven implicit conditioning, where rendered hand meshes or human poses are used to guide video generation~\cite{ye2023affordance,Ma_He_Cun_Wang_Chen_Li_Chen_2024}. These directions offer valuable insights: hand-aware supervision and keypoint-based actions are important for dexterous world modeling, while egocentric hand-motion sequences combined with scene rendering can drive plausible interaction generation, and joint-space and fingertip-space representations exhibit different trade-offs in human-to-robot transfer. However, the above conditioning strategies often compress the entire action vector into a single embedding, which inevitably collapses heterogeneous semantics, leading to poor action following and unstable physical predictions. We argue that this aggregation fundamentally breaks down in the high-DoF regime. \textbf{(3) Limited semantic grounding.} Beyond the action condition itself, recent 2D and 3D representation encoders and foundation models have shown strong potential for visual generation and spatial understanding. Recent works~\cite{sun2026vggt,zhou2024dino,zhou2025omniworld} further explore using pretrained visual or geometric features as extra predictive states. However, most action-conditioned world foundation models still rely on a single VLM reasoner to provide limited semantic information~\cite{agarwal2025cosmos}. These semantic and geometric priors remain underexplored in egocentric dexterous world models, where high-DoF hand motions require fine-grained action-spatial alignment.

To address these limitations, we propose \textbf{DexAC-WM}, an egocentric action-conditioned world model that explicitly rethinks how actions are represented and injected under high-DoF dexterous control. Our key insight is that action dimensions should remain structured throughout the conditioning process, rather than being collapsed into a single representation. Based on this principle, we introduce a \textbf{Structured Action Representation}, which decomposes high-DoF actions into semantically independent tokens with magnitude alignment, enabling balanced learning across heterogeneous motion factors. Building on this structured representation, 
we further propose a \textbf{Unified Local-Global Conditioning} module that decouples action injection into two complementary pathways. The \emph{local refinement} injects fine-grained action tokens via cross-attention to align subtle motions with local visual dynamics, while the \emph{global modulation} summarizes action intent and injects it through adaptive modulation, ensuring temporally coherent motion. This design captures both fine-grained dexterous control and global dynamics. To further enhance semantic understanding, we introduce \textbf{semantic condition} to fuse DINOv3~\cite{simeoni2025dinov3} features with VLM text embeddings via cross-addition injection.
% we further propose a \textbf{Unified Local-Global Conditionning} module that decouples conditioning into two complementary pathways. A \emph{local refinement branch} injects fine-grained action tokens into latent visual representations via cross-attention, enabling precise alignment between subtle action variations and local visual dynamics. In parallel, a \emph{global modulation branch} summarizes overall action intent and injects it through adaptive feature modulation, ensuring temporally coherent and physically consistent motion across the sequence. This unified design allows the model to capture both fine-grained dexterous control and long-range motion structure, overcoming the limitations of conventional global conditioning strategies. To further encourage model focusing on semantic-rich geometric representation, we fuse DINOv3 ~\cite{simeoni2025dinov3} feature latent output with VLM text embedding via cross addition injection.
We evaluate DexAC-WM on two large-scale egocentric datasets, where full model consistently improves both visual fidelity and action consistency over strong baselines. Our results suggest that \textbf{structured action representation and semantic conditioning are essential for scaling world models to high-DoF control regimes}, providing a new perspective on action modeling in embodied generative systems.
We summarize the main contributions as follows:
\begin{itemize}
\item We identify a core limitation of existing action-conditioned world models in high-DoF dexterous settings: global action aggregation suffers from scale imbalance across heterogeneous action dimensions, suppressing subtle but semantically critical finger signals.
\item We propose \textbf{DexAC}, a structured action-conditioning module tailored to high-DoF dexterous world models. DexAC preserves dimension-level action semantics through tokenization and re-couples heterogeneous actions with a unified local-global conditioning mechanism for fine-grained and globally coherent motion.
\item We introduce a \textbf{semantic conditioning branch} with DINOv3, which provides strong semantic priors for visual-temporal prediction, while DexAC aligns these priors to further improve action-following consistency.
\item Extensive experiments on EgoDex and EgoVerse show that DexAC with semantic condition substantially improves FID, FVD, and PCK over advanced baselines, including Wan, IRASim, and Cosmos. Further results on IRASim highlight the scalability of structured action conditioning.
\end{itemize}

\section{Related Work}

\subsection{Action-Conditioned World Models}

Recent advances in generative modeling have led to the emergence of action-conditioned world models, which simulate future observations under explicit control signals. Early works such as UniSim~\cite{yang2023learning}, Genie~\cite{bruce2024genie}, and RoboDreamer~\cite{zhou2024robodreamer} demonstrate that incorporating actions into diffusion-based video models enables interactive environment generation. Building on this idea, subsequent research explores data-driven world modeling for control, spanning both pixel-space video generation~\cite{zhu2025unified, zhu2025irasim, wang2026eva} and latent dynamics modeling~\cite{zheng2025flare, nematollahi2025lumos, escoriza2025multi, ning2025prompting}. In parallel, another line of work focuses on latent world models for long-horizon reasoning and planning, emphasizing efficiency and scalability~\cite{assran2025v,hafner2023mastering,hansen2023td,schrittwieser2020mastering}.
\begin{figure*}[t]
    \centering
    \includegraphics[width=\textwidth]{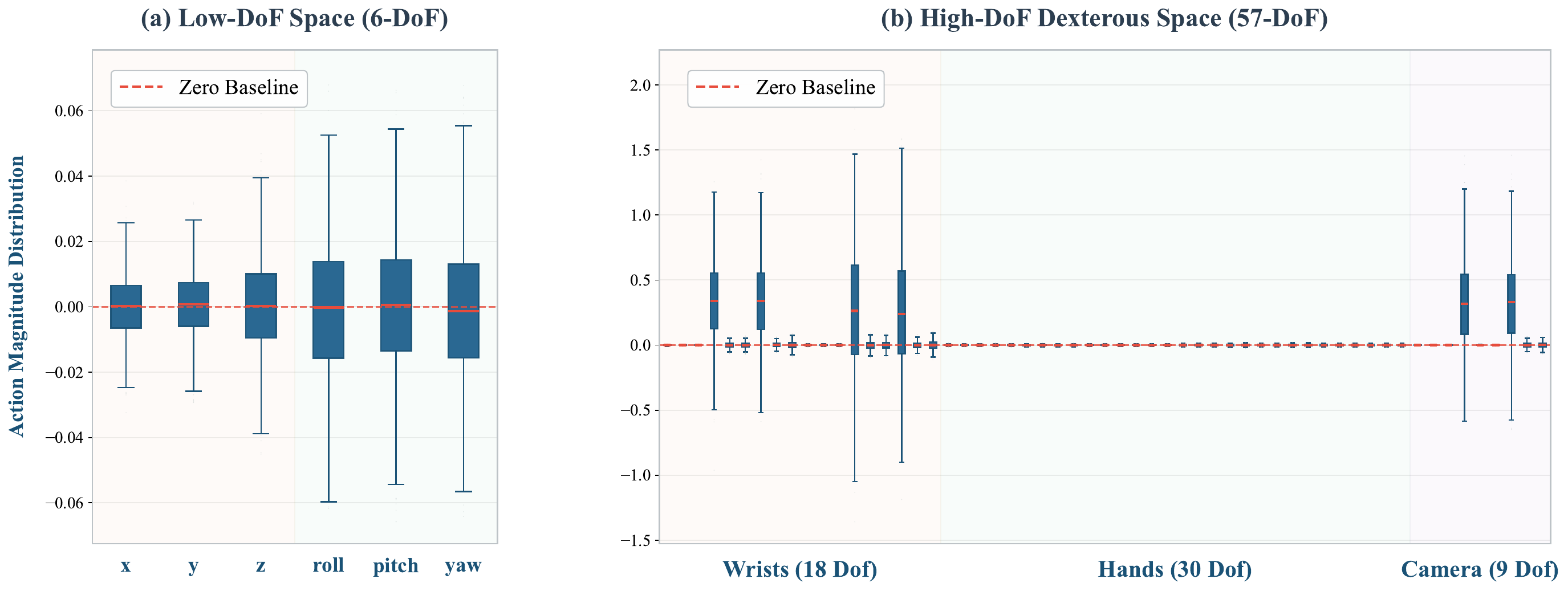}
    \caption{Empirical comparison of action magnitude distributions. \textbf{(a) Low-DoF action space (6-DoF).} The translation and rotation dimensions maintain bounded variances spanning the $10^{-2}$ scale, effectively preventing gradient domination.
    \textbf{(b) High-DoF dexterous action space (57-DoF).} The severe $10^5$ scale gap between macro-movements (wrist/camera) and micro-movements (fingers) creates critical optimization bottlenecks. Note that the finger articulations (30 DoF) appear almost flat due to their $10^{-5}$ scale compared to the $10^0$ macro-movements.
    }
    \label{fig:magnitude_comparison}
\end{figure*}
Across these directions, a common design choice is to represent actions as compact vectors and integrate them into sequence models through simple conditioning mechanisms (e.g., AdaLN or MLP concatenation)~\cite{peebles2023scalable,chen2021decision}. This formulation has proven effective in low-DoF control settings and has driven much of the recent progress in generative world modeling. At the same time, recent works increasingly explore richer visual representations, moving beyond raw pixels toward semantically meaningful feature spaces(such as pre-trained visual foundation models)~\cite{xiao2022masked,nair2022r3m,ma2022vip} to improve predictive modeling.
As action spaces become more complex, designing more expressive and structured interfaces between action and perception is becoming an increasingly important direction.

\subsection{Egocentric Human Video}

Egocentric human video has become an important data source for learning interaction dynamics, providing direct observations of human intention and object manipulation from a first-person perspective. \begin{wrapfigure}{r}{0.45\textwidth}
    \vspace{-10pt}
    \centering
    \includegraphics[width=0.43\textwidth]{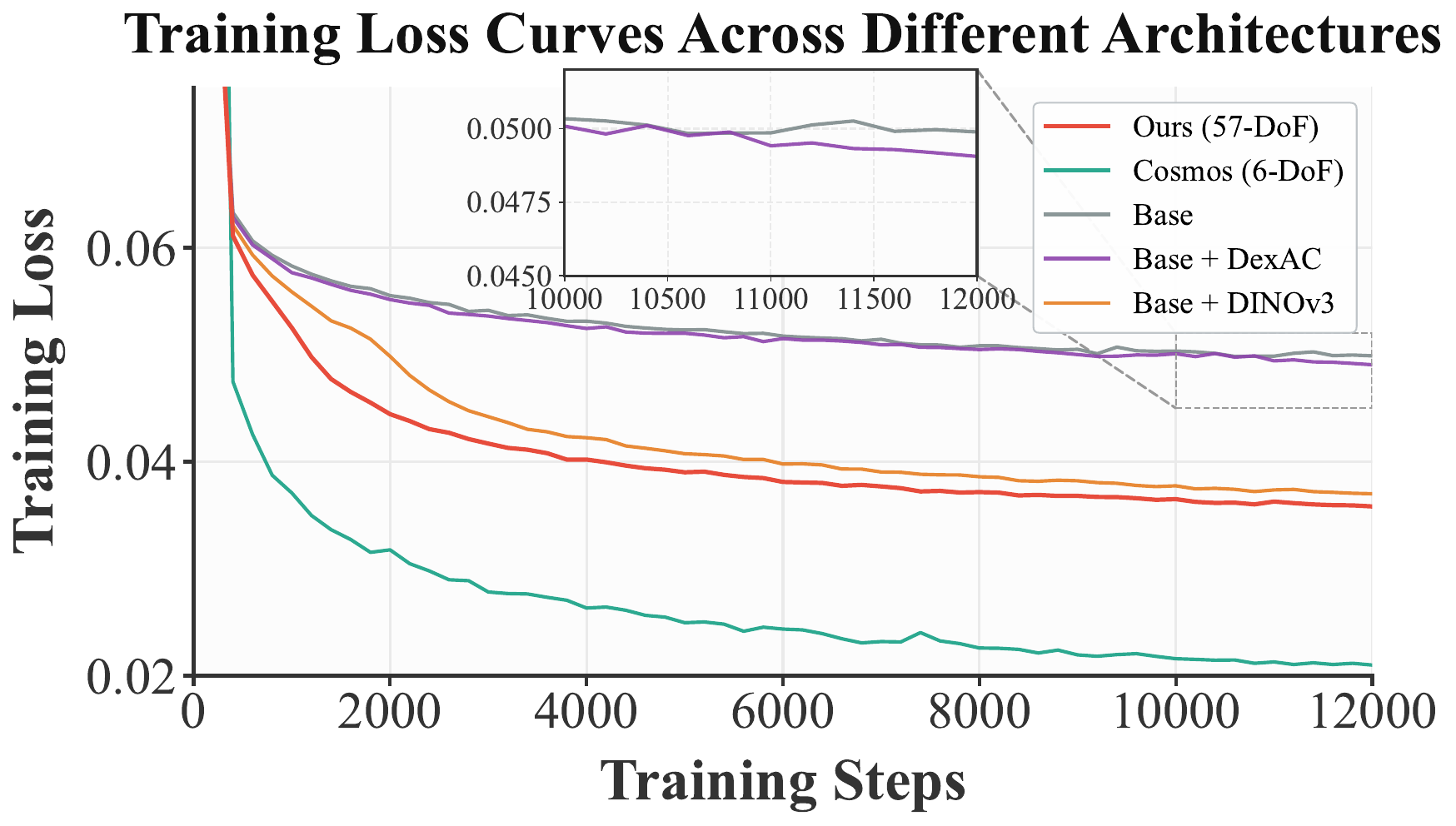}
    \caption{Training loss comparison. While the vanilla global conditioning struggles to converge in the high-DoF (57-DoF) space compared to the 6-DoF baseline, our proposed DexAC with semantic condition effectively stabilizes training and achieves a significantly lower loss.}
    \label{fig:loss_compare}
    \vspace{-18pt}
\end{wrapfigure}Large-scale datasets such as Ego4D~\cite{grauman2022ego4d}, EPIC-KITCHENS~\cite{damen2018scaling}, and HOI4D~\cite{liu2022hoi4d} have enabled significant progress in learning action representations and manipulation priors from human demonstrations. Building on these resources, prior work
has explored the extraction of policies
and transferable visual representations for downstream real-world robotic control~\cite{shaw2023videodex, yu2025egosim, goswami2026worldmodelslearningdexterous}.

A parallel line of effort has focused on constructing dedicated benchmarks for fine-grained dexterous manipulation at scale. HOT3D~\cite{banerjee2025hot3d} introduces multi-view egocentric recordings with motion-capture-level 3D hand and object poses. EgoMimic~\cite{kareer2025egomimic} pairs egocentric video with 3D hand tracking to enable co-training from human and robot data. EgoDex~\cite{hoque2025egodex} provides 829 hours of full-finger 3D hand tracking across 194 manipulation tasks collected via Apple Vision Pro. OpenEgo~\cite{jawaid2025openego} consolidates six public datasets into 1,107 hours with unified MANO annotations and intention-aligned action primitives. At a larger scale, EgoVerse~\cite{punamiya2026egoverse} spans 1,362 hours across 1,965 tasks and 240 scenes under a collaborative collection framework, while EgoScale~\cite{zheng2026egoscale} scales to over 20,000 hours, uncovering a log-linear scaling law between human data volume and downstream robot performance.

Beyond policy learning, egocentric data have increasingly been applied to predictive modeling: from navigation world models~\cite{bar2025navigation} and whole-body motion modeling~\cite{bai2025whole} to dexterous interaction modeling from human videos~\cite{goswami2026worldmodelslearningdexterous, kim2025dexterous}. This progression reflects a broader shift from coarse, low-DoF behaviors toward increasingly fine-grained, high-DoF interaction modeling, calling for representations that can capture the multi-scale and structured nature of egocentric interactions within generative world models.

\section{Case Study}
\label{sec:case_study}
\noindent \textbf{Why Vanilla Action Injection Fails in High-DoF Dexterity?}
To understand the challenges of high-DoF dexterous action conditioning, we first compare the magnitude distributions of low-DoF (6-DoF) and high-DoF (57-DoF) action spaces. As shown in Figure~\ref{fig:magnitude_comparison}, (a) 6-DoF gripper actions exhibit relatively balanced magnitudes, while (b) 57-DoF dexterous actions show strong scale heterogeneity. Large-scale motions, such as wrist and head movements, lie around the $10^0$ scale, whereas subtle finger articulations can fall to the $10^{-5}$ scale. This indicates that high-DoF dexterous actions are not merely higher-dimensional versions of low-DoF actions, but contain heterogeneous motion factors with substantially different numerical scales and semantic roles. Such scale heterogeneity directly affects the optimization dynamics of vanilla global action injection. In a standard MLP-based action encoder, all action dimensions are projected into a single shared embedding before interacting with visual features. During this process, large-magnitude wrist and head dimensions contribute more strongly to the shared action representation, while low-magnitude finger dimensions are easily weakened by dominant global motion factors. Consequently, the model shows limited sensitivity to heterogeneous action scales and subtle action-induced visual changes. \par
Action scaling or per-dimension normalization can reduce the raw numerical imbalance, but it does not solve the representation collapse caused by global aggregation. Even after scaling, a vanilla MLP still compresses wrist pose, finger articulation, and ego motion into one shared embedding, mixing semantically different motion factors before visual-action alignment. Therefore, scale alignment is necessary but insufficient for high-DoF dexterous conditioning. Table.~\ref{tab:Norm_ablation} and~\ref{tab:action_pck} further demonstrate this claim. Empirically, this optimization gap is reflected in the training loss curves in Figure~\ref{fig:loss_compare}. The same MLP-based action injection converges effectively in the 6-DoF setting, indicating that the network capacity is sufficient to represent low-DoF actions when the action dimensions are relatively balanced. However, when extended to 57-DoF dexterous actions, the flat action embedder converges more slowly and yields a higher training loss, showing that global action injection struggles to learn heterogeneous high-DoF actions. Compared with the base model, DexAC achieves a similar or slightly lower loss curve, suggesting that structured action representation alleviates the action-semantic collapse caused by scale heterogeneity to some extent. More importantly, when DexAC is combined with semantic conditioning, the model achieves lower and more stable convergence than using DexAC or semantic conditioning alone. This indicates that DexAC and semantic conditioning provide a more effective interface for aligning visual semantics with high-DoF dexterous actions.

\section{Methodology}
\begin{figure*}[t]
    \centering
    \includegraphics[width=\textwidth]{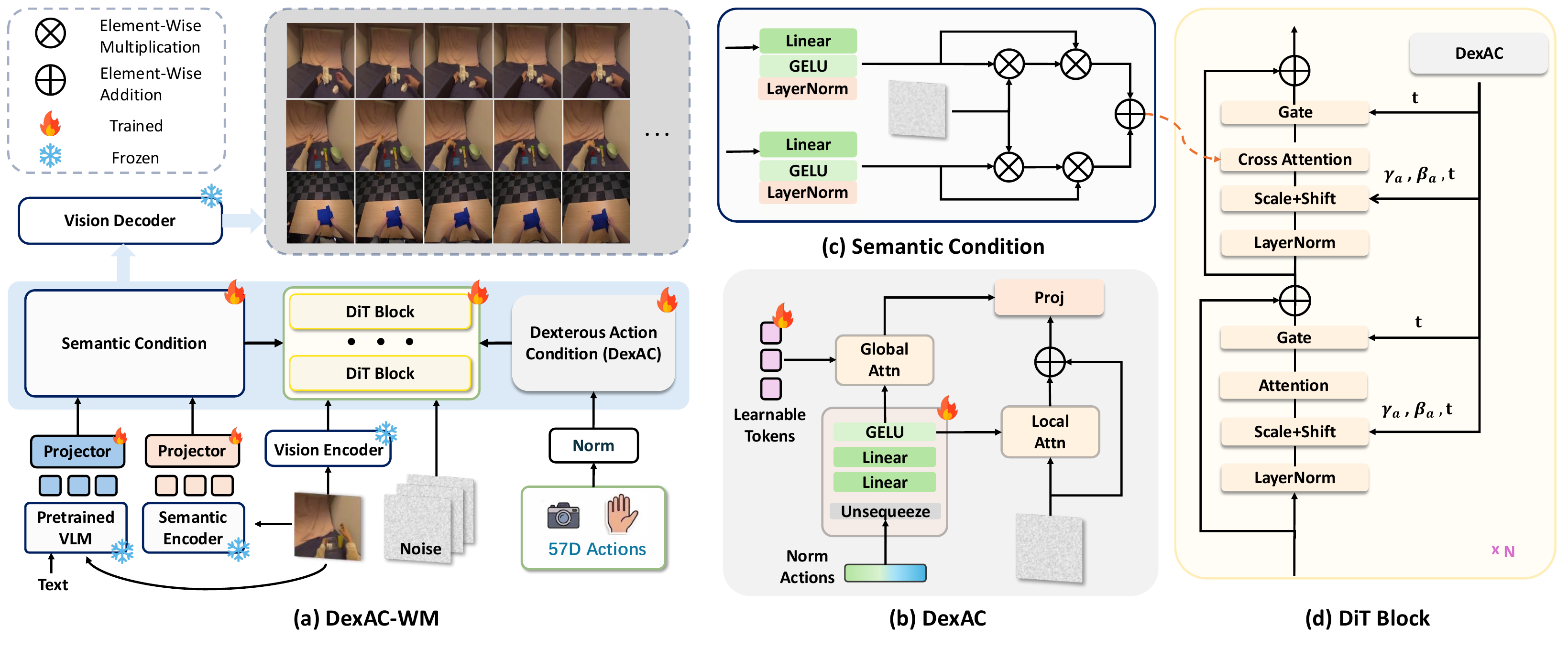}
    \caption{Diagram of DexAC-WM architecture. DexAC is designed to explicitly capture both precise local dexterity and globally coherent motion in high-DoF action regimes, while the semantic condition provides rich scene- and object-based representations for semantic understanding. (b) presents the structure of DexAC to preserve dimension-wise structured actions with local and global attention refinement for adaptive action injection. (c) introduces two vanilla cross-attention to combine DINOv3 features with VLM language embeddings for final addition. (d) presents DiT backbone architecture with 28 blocks in total, and structured action representations from DexAC are injected into each block through Adaptive Layer Normalization (AdaLN). }
        \label{fig:method_main_2}
        \vspace{-10pt}
\end{figure*}
In this section, we introduce DexAC. Its core idea is to model action conditioning as a structured process jointly optimized with visual perception, rather than a single global aggregation. We begin with the preliminary in Sec.~\ref{sec:preliminary}, followed by the framework overview in Sec.~\ref{overview}. We then present the Structured Action Representation, Unified Local-Global Conditioning, and Semantic Condition with Dual Cross Attention in Sec.~\ref{sar}, Sec.~\ref{cond}, and Sec.~\ref{condition}, respectively.

% \begin{figure}[t]
%     \centering\includegraphics[width=\textwidth]{1_cropped.pdf}
%     % \vspace{-10pt}
%     \caption{} 
%     \label{fig:photo_compare_1}
% \end{figure}
% \begin{figure}[t]
%     \centering\includegraphics[width=\textwidth]{2_cropped.pdf}
%     % \vspace{-10pt}
%     \caption{} 
%     \label{fig:photo_compare_2}
% \end{figure}
% \begin{figure}[t]
%     \centering\includegraphics[width=\textwidth]{3_cropped.pdf}
%     % \vspace{-10pt}
%     \caption{} 
%     \label{fig:photo_compare_3}
% \end{figure}
% \begin{figure}[t]
%     \centering\includegraphics[width=\textwidth]{4_cropped.pdf}
%     % \vspace{-10pt}
%     \caption{} 
%     \label{fig:photo_compare_4}
% \end{figure}

\subsection{Preliminary}
\label{sec:preliminary}
% \textbf{Problem Formulation:} given an initial visual observation $I_0$ and a sequence of actions $A \in \mathbb{R}^{T \times D_a}$,
%  where $D_a=57$ denotes the action dimension and $T$ is the action horizon, our goal is to predict future visual states regressively according to a predefined action chunk size conditioned on both the visual context and the action sequence.
% Formally, let:
% \begin{equation}
%     V^T=\{I_0,I_1,\dots,I_T\}, \quad I_t\in\mathbb{R}^{H\times W\times 3}
% \end{equation}
% denote a continuous egocentric video stream. We treat each frame $I_t$ as the visual state $S_t$, and let $A_t$ denote the action executed between $S_t$ and $S_{t+1}$. Our objective is to learn an action-conditioned transition model that predicts the next state as:
% \begin{equation}
%     \hat{S}_{t+1}=f_\theta(S_t,A_t)
% \end{equation}
% The predicted states are expected to preserve both temporal coherence and physical consistency with the underlying dexterous actions. Unlike prior gripper-based settings that typically rely on low-dimensional 6-DoF actions, our setting involves high-dimensional dexterous action signals. This fine-grained action space makes dynamics modeling substantially more challenging, and therefore requires a more structured conditioning mechanism to capture the correspondence between action.

\textbf{Problem Formulation:}
Given an initial visual observation $I_0$ and an action sequence, our
goal is to predict the future visual sequence conditioned on both the
initial visual context and the action sequence. Formally, the prediction
process is defined as:
\begin{equation}
    \hat{V}_{1:T}=f_\theta(I_0, A), \qquad A \in \mathbb{R}^{T \times D_a},
\end{equation}
where $\hat{V}_{1:T}=\{\hat{I}_1,\hat{I}_2,\dots,\hat{I}_T\}$ denotes
the predicted future visual frames, each frame
$\hat{I}_t \in \mathbb{R}^{H \times W \times 3}$, $D_a=57$ denotes the
action dimension, and $T$ is the action horizon. For step-wise modeling,
we treat each frame $I_t$ as the visual state $S_t$, and let $A_t$
denote the action executed between $S_t$ and $S_{t+1}$. The overall
sequence prediction can thus be viewed as a composition of
action-conditioned state transitions:
\begin{equation}
    \hat{S}_{t+1}=f_\theta(S_t, A_t).
\end{equation}
Compared with prior gripper-based settings that typically adopt
6-DoF actions, our setting involves 57-dimensional
dexterous action signals. Such high-DoF and fine-grained action
inputs substantially increase the complexity of visual dynamics
modeling, thereby motivating a more structured conditioning mechanism.

\noindent \textbf{Training Objective.}
To train the action-conditioned world model, we adopt a velocity prediction objective following rectified flow formulation as follows:
\begin{equation}
\mathcal{L}(\theta)
=
\mathbb{E}_{x_\tau, v, c}
\left[
\| u(x_\tau, \tau, c; \theta) - v \|_2^2
\right],
\end{equation}

\noindent where $x_\tau=(1-\tau)S_{t+1}+\tau\epsilon$ denotes the interpolated noisy latent at timestep $\tau \in [0,1]$, 
with $\epsilon\sim\mathcal{N}(0,I)$.
The target velocity is defined as $v=\epsilon-S_{t+1}$.
$c=\{c_v(S_t),\,c_a(A_t)\}$ denotes the joint conditioning, where $c_v(S_t)$ is the visual condition extracted from the current state $S_t$, and $c_a(A_t)$ is the structured action condition derived from $A_t$ via dimension-wise tokenization and unified local-global conditioning.
$u(\cdot;\theta)$ is the velocity prediction network parameterized by $\theta$.

% \noindent
% \textbf{Action imbalance issue:} XXX XXX

\subsection{Overview}
\label{overview}
% Our \textbf{DexAC} is built upon the \textit{Cosmos-Predict2.5} ~\cite{ali2025world} backbone. For text conditioning, we adopt the \textit{Cosmos-Reason1} ~\cite{azzolini2025cosmos}, which aggregates token features from multiple Transformer blocks and projects the concatenated activations into a 2048-dimensional embedding space. For visual conditioning, each RGB frame is encoded by an initialized and frozen \textit{Wan2.1}~\cite{wan2025wan} visual encoder to extract spatial contextual features. On top of this backbone, we introduce a structured action-conditioning framework tailored for high-Dof dexterous motion. 
Our \textbf{DexAC-WM} is built upon the \textit{Cosmos-Predict2.5}~\cite{ali2025world} backbone. For text conditioning, we adopt \textit{Cosmos-Reason1}~\cite{azzolini2025cosmos}, which aggregates multi-layer Transformer features into a 2048-dimensional embedding. For visual conditioning, each RGB frame is encoded by a frozen \textit{Wan2.1}~\cite{wan2025wan} visual encoder to extract spatial features. Our action representation module consists of three key components. 1) \textbf{Structured Action Representation}. Instead of flattening the action sequence into a single embedding for injection, we introduce a structured action tokenizer (SAT) to produce a sequence of per-dimension action tokens, preserving the temporal evolution and semantic independence of different motion factors. 
% 2) A \textit{Unified Local-Global Conditioning} module, which combines both \textit{local action refinement} and \textit{global action modulation}. The local action refinement branch enhances the sensitivity of latent noise tokens to fine-grained local action changes, allowing subtle dexterous motions to be better preserved during generation. The global action modulation branch uses a learnable action query to summarize the action tokens into a compact global context, providing consistent sequence-level action guidance to the backbone. 
\textbf{2) Unified Local-Global Conditioning}. 
We combine local refinement and global modulation in a unified pipeline. The local branch injects fine-grained action tokens to capture subtle motion changes, while the global branch summarizes action tokens via a learnable query to provide coherent sequence-level guidance.
\textbf{3) Semantic Condition with Dual Cross-Attention}.
To enhance the semantic representation of the model, we introduce a semantic condition mechanism that jointly inject DINOv3 features and text embeddings into the latent space, where latent tokens act as queries and multi-modal features serve as keys and values. By aggregating them through cross-attention, the model jointly reasons over geometry and intent, improving spatio-temporal consistency in generation.

% 3) \textbf{\textit{Action Adaptative Injection}}. To better align the structured action conditioning with the flow-matching objective, we introduce an adaptive action projection that fuses the global modulation signal and the locally refined action features, and injects them into the backbone through Adaptive Layer Normalization (AdaLN). This design allows the model to jointly capture fine-grained action-following dynamics and globally coherent motion trends. 

% Finally, the conditioned latent representations are optimized under the \textit{flow-matching} objective to predict denoised latent video frames. Through this unified pipeline, the model is able to generate temporally coherent and physically consistent future observations conditioned on high-Dof dexterous action sequences.

\subsection{Structured Action Representation}
\label{sar}
To model high-DoF dexterous manipulation under egocentric observations, we design a structured action tokenizer that explicitly captures the relative motion of both wrists, finger articulations, and head ego-motion. Unlike prior works such as DexWM ~\cite{goswami2026worldmodelslearningdexterous}, which represent hand motion using camera-centric keypoint displacements and Euler-angle pose deltas, our method adopts a physically grounded rigid-body formulation that is more consistent with the underlying hand-object interaction dynamics.
\textbf{\textit{Action Representation.}} 
% Specifically, given the hand pose at two consecutive time steps $t$ and $t+1$, we compute the relative transformation as $T_t^{-1} T_{t+1}$, where $T = [R, p] \in SE(3)$ denotes the hand pose, with $R \in SO(3)$ representing rotation and $p \in \mathbb{R}^3$ representing translation. Based on this formulation, the relative translation is defined as $R_t^{\top}(p_{t+1}-p_t)$ and the relative rotation is given by $R_t^{\top} R_{t+1}$.
% To avoid the discontinuity and ambiguity of Euler angles~\cite{zhou2019continuity}, we represent rotation using the continuous 6D rotation parameterization.
% \noindent Our final action vector consists of three components: (1) relative wrist motion for both hands, including relative translation and rotation. (2) Finger joint displacements that describe fine-grained hand articulation and (3) Head rotation changes that capture ego-motion of the camera wearer. This yields a unified high-Dof action representation that jointly models global hand movement, local finger motion, and viewpoint variation,
% which provides geometric consistency across time and improves robustness under egocentric camera motion. Compared with camera-centric displacement-based representations, our formulation better decouples hand motion from viewpoint changes, leading to more stable learning and more physically consistent action-conditioned prediction.
Specifically, given hand poses $T_t, T_{t+1} \in SE(3)$ at consecutive times, the relative transformation is $T_t^{-1} T_{t+1}$, yielding relative translation $R_t^\top(p_{t+1}-p_t)$ and relative rotation $R_t^\top R_{t+1}$. To avoid Euler angle issues, we adopt the continuous 6D rotation representation. The final action vector comprises: (1) relative wrist motion (translation + rotation) for both hands, (2) finger joint displacements for fine articulation, and (3) head rotation changes for camera ego-motion. This unified high-DoF representation jointly models global hand movement, local finger motion, and viewpoint variation, ensuring geometric consistency and robustness under egocentric motion. Unlike camera-centric displacements, it decouples hand motion from viewpoint changes, enabling more stable learning and physically consistent action-conditioned prediction.

\textbf{\textit{Structured Action Tokenizer (SAT).}}
%由于动作维度的异构，手和腕部的强度和变化往往呈现不同的分布，因此，我们首先针对每个的动作维度进行独立归一化让网络更好的学习和区别不同强度变化的动作表征。不同于Cosmos-predict2.5，直接将所有动作维度信息展开成一个全局向量（B，D），这极大程度的损失了在时间维度上不同动作信息的具体来说，我们首先计算不同动作维度的均值/miu和方差/sigma，
Due to the heterogeneous nature of dexterous actions, different components (e.g., fingers and wrists) exhibit distinct scales and statistical distributions. To facilitate stable learning and better distinguish action dynamics of varying magnitudes, we first normalize each action dimension independently. Given an action sequence $a \in \mathbb{R}^{T \times D_a}$, we compute the mean $\mu_i$ and standard deviation $\sigma_i$ for each action dimension $i$ over the training dataset. The normalized action is then obtained:
\begin{equation}
    \tilde{a}_{t,i} = \frac{a_{t,i} - \mu_i}{\sigma_i + \epsilon}
\end{equation}
where $\epsilon$ is a small constant for numerical stability. Our dimension-wise normalization preserves the statistical characteristics of each action component and provides a more unique representation for downstream conditioning.
Given an action sequence of $(B, T, D_{a})$, we first preserve its original temporal and dimensional structure rather than flattening it into a single vector. Each action entry is independently projected into a latent token space, producing per-timestep, per-dimension features:
\begin{equation}
h_{t,i} = \text{Linear}(\tilde{a}_{t,i}), \qquad h_{t,i} \in \mathbb{R}^{d}
\end{equation}
We then perform learnable temporal fusion along the horizon dimension to aggregate the dynamics of each action dimension into a single token. This results in a structured token set of shape $(B, N, C)$, where N presents each token corresponds to one action dimension and summarizes its temporal evolution. To aggregate temporal dynamics, we apply a learnable linear fusion along the temporal dimension followed by a GELU activation:
\begin{equation}
A_{\mathrm{tok}} = \text{GELU}(\text{Linear}([h_{1,i}, h_{2,i}, \dots, h_{T,i}])\big)
\end{equation}
This yields a set of dimension-wise action tokens $A_{\mathrm{tok}} \in \mathbb{R}^{d}$, where each token $A_{\mathrm{tok}}$ summarizes the temporal evolution of the $i$-th action dimension. Compared with conventional MLP-based action embedding, which compresses the entire action sequence into one global representation, this design preserves dimension-level semantics and provides a more suitable basis for fine-grained action conditioning.
% Unlike Cosmos-Predict2.5 \cite{}, which flatten all action dimensions into a single global vector, 
% we eager to preserve maximum action representation for each dimension. Specifically, we process the action sequence in a dimension-wise manner while preserving its temporal structure. Each action entry $a_{t,i}$ is first normalized by mean and variance to promote the learning process, and then independently projected into a latent space via a linear layer:

% Compared with using a single global embedding for the entire action sequence, this design preserves dimension-level semantics and provides a more suitable basis for fine-grained action conditioning.
% this design preserves fine-grained action structure and facilitates more precise downstream conditioning.

\subsection{Unified Local-Global Conditioning}
\label{cond}
To effectively inject high-DoF dexterous actions into the video generation backbone, we propose a \textit{Unified Local-Global Conditioning}, which integrates both \textit{local action refinement} and \textit{global action modulation} into a unified conditioning pipeline. Unlike prior action injection strategies \cite{ali2025world} that directly compress the entire action vector into a single embedding through a simple MLP, our design preserves the structural heterogeneity of different action dimensions and allows the model to selectively emphasize important motion components. Existing action-conditioned world models often aggregate all action dimensions into one global representation, implicitly averaging the contributions of heterogeneous motion factors. While this strategy is simple, it tends to weaken the representation of critical yet small-magnitude action dimensions, especially in dexterous manipulation, where subtle finger motions and ego-motion changes be easily overwhelmed by larger-scale wrist motion. To address this limitation, we design a two-stage conditioning mechanism: a \textit{local action refinement} branch that directly adjusts latent noise according to fine-grained action tokens, and a \textit{global action modulation} branch that summarizes the overall action intent and injects it into the DiT through adaptive feature modulation.

\textbf{\textit{Local Action Refinement.}}
Since the raw action values are numerically very small, we multiply each dimension by a constant scaling factor before tokenization to enhance conditioning strength and improve optimization stability. Each rescaled action dimension is then projected into the latent embedding space through a lightweight tokenizer.
To enable direct interaction between action and latent visual dynamics, we introduce a linear projection layer that maps the action tokens into the latent feature space used by the diffusion backbone. Let the noise tokens $Z \in \mathbb{R}^{N \times d}$, where $N$ is the number of latent tokens. After projection, we perform local action injection through local attention:
\begin{equation}
Z^{\mathrm{local}} = Z + \mathrm{Attn}(Q=Z,\; K={A}_{\mathrm{tok}},\; V={A}_{\mathrm{tok}})
\end{equation}
where ${A}_{\mathrm{tok}}$ denotes the projected action tokens. This module allows each latent token to actively query the structured action representation, so that fine-grained action information can directly refine the latent noise distribution. The residual connection preserves the original latent structure while enabling action-conditioned local modulation. 

% This is particularly important in our setting, since the motion statistics across action dimensions are highly imbalanced: many fingertip and camera transition components exhibit only very small variation, making them difficult to learn when mixed with larger-scale motion signals.

\textbf{\textit{Global Action Modulation.}}
While local attention provides fine-grained action priors for noise, it operates in a token-wise manner and lacks an explicit mechanism to capture the global structure of the action sequence. As a result, the model struggles to maintain coherent motion trends across frames, especially in high-DoF dexterous settings where actions exhibit strong temporal and structural dependencies. To preserve the model’s ability to perceive structured actions at a holistic level, we further introduce a \textit{global action modulation} branch. Specifically, we define a learnable global query $q_g \in \mathbb{R}^{1 \times d}$, which is used to summarize the action tokens:
\begin{equation}
A_{g} = \mathrm{Attn}(Q=q_g,\; K=A_{\mathrm{tok}},\; V=A_{\mathrm{tok}})
\end{equation}
where $A_g \in \mathbb{R}^{1 \times d}$ is the global action context. This branch extracts a compact representation that captures the overall motion intent of the current action chunk. We then convert the global action context into scale and shift parameters:
\begin{equation}
(\gamma_a,\beta_a)=f_{\mathrm{AdaLN}}(A_{g})
\end{equation}
where $\gamma_a,\beta_a \in \mathbb{R}^{d}$. These parameters are used to modulate the locally refined latent features through Adaptive LayerNorm:

\begin{equation}
\mathrm{AdaLN}(Z^{\mathrm{local}};\gamma_a,\beta_a)=\gamma_a \odot \mathrm{LN}(Z^{\mathrm{local}})+\beta_a.
\end{equation}

By doing so, the local action-conditioned latent features are further aligned with the global action context. This modulation improves the consistency of generated dynamics at the sequence level and provides a stronger global control signal to the DiT backbone.

\textbf{\textit{Action Adaptive Injection.}}
Finally, we combine the local and global branches into a adaptive layer normalization module for each DiT block. The local cross-attention branch ensures that latent noise tokens can directly perceive structured action information, enabling fine-grained action following. The global modulation branch summarizes the action tokens into a compact context vector and injects it through AdaLN, ensuring that the overall motion pattern remains coherent and physically consistent.

Formally, the fused latent representation is written as:
\begin{equation}
Z^{\mathrm{fused}} = \mathrm{AdaLN}(Z^{\mathrm{local}};\gamma_a,\beta_a).
\end{equation}

\noindent Compared with conventional single-embedding action injection, our design avoids averaging all action dimensions too early. Instead, it preserves dimension-level structure, emphasizes subtle but semantically important motion signals, and combines local refinement with global modulation in a complementary manner. As a result, the model better captures both fine-grained dexterous action details and long-range motion consistency during generation.
% \begin{wrapfigure}{r}{0.42\textwidth}
%     \vspace{-10pt}
%     \centering
%     \includegraphics[width=0.40\textwidth]{figure/method_1.pdf}
%     \caption{The structure of semantic condition. We extend vanilla self-attention to explicitly unify semantic-rich and VLM embedding features to enhance the spatial perception capability with strong semantic priors.}
%     \label{fig:method_1}
%     \vspace{-18pt}
% \end{wrapfigure}
\subsection{Semantic Condition with Dual Cross-Attention}
\label{condition}
To better capture the semantics of hand-object interaction, we introduce visual features with strong 2D structured geometric priors to complement the ViT encoder, which is semantically expressive but relatively weak in modeling fine-grained geometry. 
These two representations are complementary: the ViT encoder provides high-level semantic understanding, while DINO features offer dense spatial representations aligned with image content, thereby strengthening the model’s ability to localize hand-object regions and reason about scene geometry. 
Inspired by~\cite{zhou2024dino,zhu2023vima,chen2025internvla}, we adopt a dual cross-attention to jointly inject DINOv3-L visual features and text embeddings into the latent space. Specifically, latent tokens serve as queries, while DINO tokens and text embeddings act as keys and values. DINO features provide dense, image-aligned spatial cues for fine-grained grounding of object geometry and scene structure, whereas text embeddings provide compact global semantics for high-level intent guidance. To balance the contributions of these two modalities, we concatenate DINO tokens and text tokens along the sequence dimension and aggregate the multi-modal context through cross-attention. This design enables the model to jointly reason over spatial details and semantic cues, thereby improving spatio-temporal consistency in both future appearance and motion generation.

\section{Experiments}
In this section, we first introduce implementation details in Sec.~\ref{sec:exp_setup}. We then present both quantitative and qualitative comparisons against a SOTA baseline in Sec.~\ref{sec:evaluation}, highlighting improvements in visual quality, temporal coherence, and action consistency. Finally, we perform detailed ablation studies in Sec.~\ref{sec:ablation} to analyze the contributions of structured action representation, unified local-global conditioning, and semantic condition with dual cross-attention and provide more details of the mechanism of our DexAC. 

\subsection{Experimental Setup}
\label{sec:exp_setup}
\textbf{\textit{Datasets.}} We evaluate our DexAC-WM on two large-scale egocentric human manipulation datasets. \textbf{EgoDex}~\cite{hoque2025egodex} comprises 829 hours of 1080p egocentric video at 30 Hz, containing 194 manipulation tasks involving 500 distinct objects, with rich multimodal annotations including 3D skeletal poses for the upper body and 25 keypoints per hand, camera intrinsics and extrinsics, and confidence scores for all pose estimates. \textbf{EgoVerse}~\cite{punamiya2026egoverse} is a collaborative egocentric dataset spanning 1,362 hours across 80k episodes, covering 1,965 tasks, 240 scenes, and 2,087 unique demonstrators, with standardized formats and manipulation-relevant annotations sourced from researchers and industry partners worldwide. For training and evaluation, we use the whole training set and test set data from EgoDex, and EgoVerse-A data from EgoVerse.

\noindent \textbf{\textit{Evaluation metrics.}} We use PSNR~\cite{hore2010image} and SSIM~\cite{wang2004image} for pixel-level and structural quality, LPIPS~\cite{zhang2018unreasonable} for perceptual similarity, and FID~\cite{heusel2017gans} and FVD~\cite{unterthiner2018towards} to evaluate spatial quality and spatio-temporal realism, respectively. Specifially, FID/FVD use the Cosmos-Cookbook protocol. We further report PCK@10 and PCK@20~\cite{yang2012articulated} on predicted keypoints to evaluate fine-grained localization accuracy. PCK@10 reflects stricter precision, while PCK@20 captures overall motion correctness.

\noindent \textbf{\textit{Baselines.}}
To comprehensively evaluate the proposed DexAC-WM, we focus on Cosmos-Predict2.5~\cite{ali2025world}, Wan-Control~\cite{wan2025wan} and IRASim~\cite{zhu2025irasim} as baselines, as they represent  strong and state-of-the-art action-conditioned world modeling frameworks under a comparable setting for fair comparison.
% \textbf{Zero-shot Video Foundation Models}: We evaluate VideoX-Fun~\cite{}, Wan2.1 ~\cite{}, and Wan2.2 ~\cite{}, which represent state-of-the-art diffusion models for general video generation. We benchmark these models in a zero-shot setting to assess their inherent physical priors and zero-shot generalizability when confronted with complex, fine-grained egocentric manipulation tasks.
% Navigation and Unified World Models : We compare against Navigation World Model (NWM) ~\cite{}, which predicts future egocentric observations conditioned on navigation actions, and Unified World Model (UWM) ~\cite{}. We adapt these models to our setting to evaluate the scalability of their latent dynamics when transferred from coarse locomotion to high-DoF dexterous manipulation.
% Action-Conditioned Foundations: We benchmark against the Cosmos-Predict series ~\cite{}, encompassing both text-conditioned and action-conditioned (AC) variants, as well as direct inference Video-to-World models. These serve as the direct foundational architecture for our proposed method.

\noindent\textbf{\textit{Implementation details.}}  We train all models on the EgoDex dataset using 8 NVIDIA H200 GPUs with each batch size of 64. We adopt a sliding-window sampling strategy with 30-frame non-overlapping windows and 2 random starting positions per window to construct 13-frame training clips, consisting of 1 conditioning frame and a 12-step prediction horizon. We use 57-DoF action vectors as conditional inputs, including 15 dimensions per hand for finger articulation, 9 dimensions per wrist for 6D rotation representation, and 9 dimensions for relative camera pose. We employ a 2B action-conditioned backbone and finetune the model using an MSE loss using AdamW optimizer with learning rate of $4 \times 10^{-5}$ and a weight decay of 0.1.

\subsection{Evaluation}
\label{sec:evaluation}
We evaluate generated videos using a fixed temporal protocol, following the Cosmos-Predict2.5-DROID setting with a 12-frame prediction chunk. Specifically, all generated videos are trimmed to a uniform duration and temporally resampled to 30 frames per second, resulting in 26-frame sequences for evaluation. This setting is also comparable to Ctrl-World~\cite{guo2025ctrl}, which evaluates 6-frame predictions at 5 Hz within a 1 second horizon. All evaluations are conducted at a spatial resolution of $224 \times 224$.

\textbf{Quantitative Results:} 
Table~\ref{tab:video_world_models}  compare the performance of our method with that advanced action-conditioned world models on the EgoDex and EgoVerse datasets. Although generative-based baselines (such as Wan2.1 and Wan2.2) obtain moderate reconstruction quality but show weak temporal coherence, with FVD exceeding 1200 on both datasets. This indicates that large-scale video priors alone are not sufficient for action-conditioned dexterous prediction. Applying DexAC to IRASim further improve action consistency, with
\begin{figure}[H]
\centering\includegraphics[width=\textwidth]{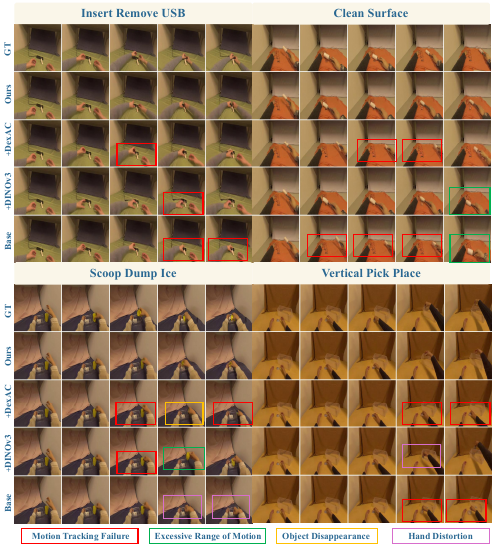}
    \vspace{-10pt}
    \caption{Qualitative comparison of action-conditioned video prediction on Egodex~\cite{hoque2025egodex}. From top to bottom, the rows show the frame sequences of the Ground Truth (GT),  our proposed method (Ours), Cosmos w/ DexAC (+DexAC), Cosmos w/ DINOv3 (+DINOv3), and Base (Cosmos).} 
    \label{fig:photo_compare}
\end{figure} \noindent PCK@10 gains of +6.18 on EgoDex and +3.0 on EgoVerse, and also reduces FVD on both datasets. These results show that DexAC provides complementary gains across different base architectures. For Cosmos-based variants, adding DINOv3 alone improves visual perception quality in FID on both EgoDex and Egoverse, but increases FVD from 352.19 to 977.68 in Egodex, suggesting weaker temporal stability. DexAC achieve the best PCK, and second-best in FVD in Egodex and second-best in PCK@20 in Egoverse, which demonstrates the gains in temporal coherence and action consistency. Our full model achieves the best FID and FVD on both benchmarks, reaching 106.6/284.40 on EgoDex, 139.60/830.03 on EgoVerse and best PCK and second-best results on Egoverse and Egodex. These results indicate that the joint effect of DINOv3 and DexAC better pre-
\begin{figure}[H]
    \centering\includegraphics[width=\textwidth]{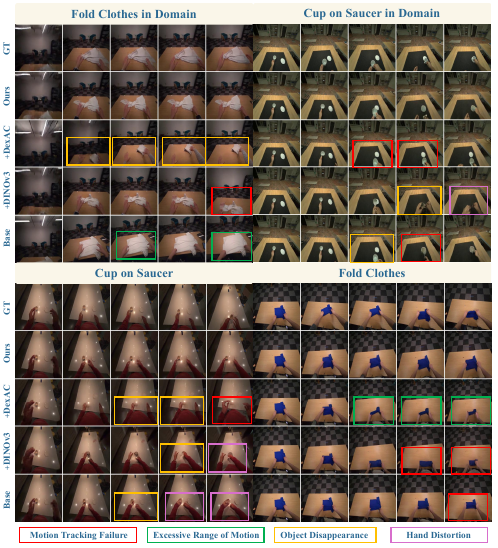}
    \vspace{-10pt}
    \caption{Qualitative comparison of action-conditioned video prediction on EgoVerse~\cite{punamiya2026egoverse}. From top to bottom, the rows show the frame sequences of the Ground Truth (GT), our proposed method (Ours), Cosmos w/ DexAC (+DexAC), Cosmos w/ DINOv3 (+DINOv3), and Base (Cosmos).} 
    \label{fig:photo_compare_egoverse}
\end{figure} \noindent serves semantic representations of structured actions, significantly improving action-temporal prediction.

% These results suggest that DINOv3 and DexAC are complementary: DINOv3 improves semantic grounding and visual fidelity, while DexAC strengthens action-aware temporal prediction.

\begin{table*}[t]
\caption{Quantitative comparison of advanced action-conditioned world models on EgoDex~\cite{hoque2025egodex} and EgoVerse~\cite{punamiya2026egoverse}.
Higher PSNR and SSIM indicate better reconstruction quality, while lower LPIPS, FID and FVD indicate better perceptual and temporal quality. Higher PCK represents better action consistency. The best and second-best AVG are highlighted in bold and underlined, respectively.}
\label{tab:video_world_models}
\centering
\renewcommand{\arraystretch}{1.4}
\resizebox{\textwidth}{!}{
\begin{tabular}{l|ccccccc}
\toprule
Baselines 
& PSNR $\uparrow$ 
& SSIM $\uparrow$ 
& LPIPS $\downarrow$ 
& FID $\downarrow$
& FVD $\downarrow$
& PCK@10 $\uparrow$
& PCK@20 $\uparrow$ \\
\midrule\midrule
\rowcolor{rowgray}
\multicolumn{8}{c}{\textbf{EgoDex}~\cite{hoque2025egodex}} \\
Wan2.1-Fun-1.3B-Control~\cite{wan2025wan}              & 21.89 & \textbf{0.89} & 0.34 & 194.11 & 1532.49 & 20.35 & 36.89 \\
Wan2.2-Fun-5B-Control~\cite{wan2025wan}              & 22.97 & 0.73 & 0.31 & 167.98 & 1434.19 & 21.51 & 36.84 \\
IRASim~\cite{zhu2025irasim}         & 22.12 & 0.80 & \underline{0.20} & 153.81 & 615.21  & 27.76 & 44.84 \\
\rowcolor{rowgray}
IRASim + DexAC       & 23.11 & \underline{0.81}& \textbf{0.16}& 142.76 & 565.30 {\scriptsize ($\downarrow$8.11\%)}  & \underline{33.94} {\scriptsize (+22.26\%)} & 51.37{\scriptsize (+14.56\%)} \\
Cosmos-Predict2.5-2B (Base)~\cite{ali2025world} 
& 25.02 & 0.80 & 0.25 & 114.51 & 352.19 & 31.07 & 58.33 \\
Base + DINOv3
& \textbf{25.74} 
& \underline{0.81} 
& 0.23
& \underline{110.25}
& 977.68 
& 33.86 
&  \underline{60.78} \\
\rowcolor{rowgray}
Base + DexAC
& \underline{25.14} {\scriptsize (+0.47\%)}
& 0.80 
& 0.25 
& 114.26 {\scriptsize ($\downarrow$0.22\%)} 
& \underline{349.29} {\scriptsize ($\downarrow$0.82\%)}
& \textbf{34.15} {\scriptsize (+9.91\%)} 
& \textbf{61.41} {\scriptsize (+5.28\%)} \\
\rowcolor{rowgray}
Ours (Base + DINOv3 + DexAC)
& 25.13 {\scriptsize (+0.44\%)} 
& 0.80 {\scriptsize (+0.57\%)} 
& 0.24 {\scriptsize ($\downarrow$1.43\%)} 
& \textbf{106.67} {\scriptsize ($\downarrow$6.84\%)}  
& \textbf{284.40} {\scriptsize ($\downarrow$19.25\%)} 
& 32.70 {\scriptsize (+5.25\%)} 
& 60.59 {\scriptsize (+3.87\%)}  \\
\midrule
\rowcolor{rowgray}
\multicolumn{8}{c}{\textbf{EgoVerse}~\cite{punamiya2026egoverse}} \\
Wan2.1-Fun-1.3B-Control~\cite{wan2025wan}              & 22.43 & 0.74 & \underline{0.37} & 176.99 & 1370.18  & 20.17 & 33.95 \\
Wan2.2-Fun-5B-Control~\cite{wan2025wan}              &  21.93     &    \textbf{0.79}  & 0.41     &  151.97      &     1203.89    &   25.74    &   41.32   \\
IRASim~\cite{zhu2025irasim}              & \underline{22.59} & 0.71 & \textbf{0.35} & 229.74 & 989.21  & \underline{41.68} & 57.90 \\
\rowcolor{rowgray}
IRASim+DexAC       &  \textbf{23.66}  {\scriptsize (+4.74\%)}     & \underline{0.75}  {\scriptsize (+5.63\%)}    & 0.39     &  224.77 {\scriptsize ($\downarrow$2.16\%)}      &   963.25 {\scriptsize ($\downarrow$2.62\%)}     &  \textbf{44.68}  {\scriptsize (+7.20\%)}   &  \underline{58.12}  {\scriptsize (+0.38\%)}   \\
Cosmos-Predict2.5-2B (Base)~\cite{ali2025world} & 21.45 & 0.63 & 0.39 & \underline{151.62} & 955.74 &  24.58     & 41.16      \\
Base + DINOv3       &  21.09      &  0.62     &  0.41     &  162.13       &     \underline{857.82}    &   26.45    &   41.73    \\
\rowcolor{rowgray}
Base + DexAC          & 21.37& 0.63 & 0.40 & 152.10 & 919.56 {\scriptsize ($\downarrow$3.79\%)} &  24.24     &  42.72  {\scriptsize (+3.79\%)}   \\
\rowcolor{rowgray}
Ours (Base + DINOv3 + DexAC) &21.67 {\scriptsize (+1.03\%)} &0.64 {\scriptsize (+1.59\%)}&0.38 {\scriptsize ($\downarrow$2.56\%)} &\textbf{139.60} {\scriptsize ($\downarrow$7.93\%)}&\textbf{830.03} {\scriptsize ($\downarrow$13.15\%)} & 40.62 {\scriptsize (+65.26\%)}&\textbf{60.51} {\scriptsize (+47.01\%)} \\
\bottomrule
\end{tabular}
}
\end{table*}

\textbf{Qualitative Results:} 
Figure~\ref{fig:photo_compare} and~\ref{fig:photo_compare_egoverse} show qualitative comparisons on EgoDex and EgoVerse. The base model can reconstruct the overall scene layout, but it often produces distorted hand poses and mismatched hand-object interactions. Under object occlusion, both the base and +DexAC show noticeable object disappearance. DINOv3 alleviates this situation by providing object- and scene-level semantic, but its predictions still exhibit motion drift, although the overall motion direction is partially preserved. DexAC better maintains the structural integrity of hand motions, but it is less consistent in modeling object interactions. For deformable object tasks (folding clothes) and multi-object scenarios, maintaining hand-object consistency becomes the main challenge. The base, +DINOv3, and +DexAC show different degrees of hand-object deformation. DexAC preserves more coherent motion trajectories and hand details, while the base and DINOv3 only exhibit more visible action drift. In comparison, the full model better preserves hand poses, motion trajectories, and hand-object interaction consistency, while reducing accumulated motion drift. These results indicate that structured action representation and action-aware semantic conditioning jointly help model both subtle dexterous motions and globally consistent interaction behavior.

\subsection{Ablation Study}

%In the ablation section, we answer the following question to validate the effectiveness of our design: a
We conduct a series of ablations to examine why DexAC and a semantic condition branch improve high-DoF dexterous world modeling. Specifically, we investigate:

\label{sec:ablation}
% We conduct multiple ablation experiments to verify the module gains of the DexAC.  \\
% \begin{table}[t]
% \centering
% \scriptsize
% \setlength{\tabcolsep}{2pt}
% \resizebox{\columnwidth}{!}{
% \begin{tabular}{l|l|ccccc}
% \toprule
% Setting & Param & PSNR $\uparrow$ & SSIM $\uparrow$ & LPIPS $\downarrow$ & FID $\downarrow$  \\
% \midrule
% \multirow{3}{*}{DINO Fusion Param}
% & 0.2 & 26.12 & 0.82  & 0.23 & 114.24  \\
% & 0.4 &  &  &  &  \\
% & 1 &  &  &  &   \\
% \bottomrule
% \end{tabular}}
% \caption{Ablation study on the DINOv3 injection strength. (train in part1 with 2005 steps)}
% \label{tab:data_scale}
% \end{table}

% \begin{table}[t]
% \centering
% \scriptsize
% \setlength{\tabcolsep}{2pt}
% \resizebox{\columnwidth}{!}{
% \begin{tabular}{l|l|ccccc}
% \toprule
% Setting & Components & PSNR $\uparrow$ & SSIM $\uparrow$ & LPIPS $\downarrow$ & FID $\downarrow$ & FVD $\downarrow$ \\
% \midrule
% \multirow{4}{*}{Action Dim Rep.}
% & only wrist (18 dim) &  &  &  &  &  \\
% & only hand (30 dim) &  &  &  &  &  \\
% & w/o cam motion (48 dim) &  &  &  &  &  \\
% & full dim (57 dim) -- ours &  &  &  &  &  \\
% \bottomrule
% \end{tabular}}
% \caption{Ablation study on different action dimension representations for heterogeneous dexterous control.}
% \label{tab:action_dim}
% \end{table}

\textbf{a. What are the roles of different components in DexAC?}
We conduct ablation studies to evaluate the contribution of each component in our structured action conditioning framework, as summarized in Table~\ref{tab:depth_ablation}. Removing either local refinement or global modulation consistently degrades performance, confirming that both branches are important for high-DoF action conditioning. In particular, removing global modulation leads to the largest drop in overall performance, with clear degradation in reconstruction quality, perceptual fidelity, temporal coherence, and action consistency, indicating that global action intent is essential for stabilizing generation. Removing local refinement also degrades performance, especially in FVD and PCK, showing that fine-grained action injection is necessary for preserving dexterous motion details. Compared with the simple MLP-based conditioning baseline, our full condition achieves the best overall results, notably improving FID from 114.51 to 106.67 and FVD from 352.19 to 284.40, while also increasing PCK@10 from 31.07 to 32.70 and PCK@20 from 58.33 to 60.59 on EgoDex. These results demonstrate that the proposed unified local-global conditioning is more effective than conventional global action embedding for high-DoF dexterous actions. Computation analysis of DexAC can refer to Table~\ref{tab:action_param_comparison} in Sec.~\ref{sec:dexac}.

\begin{figure}[t]
    \centering
    \includegraphics[width=\linewidth]{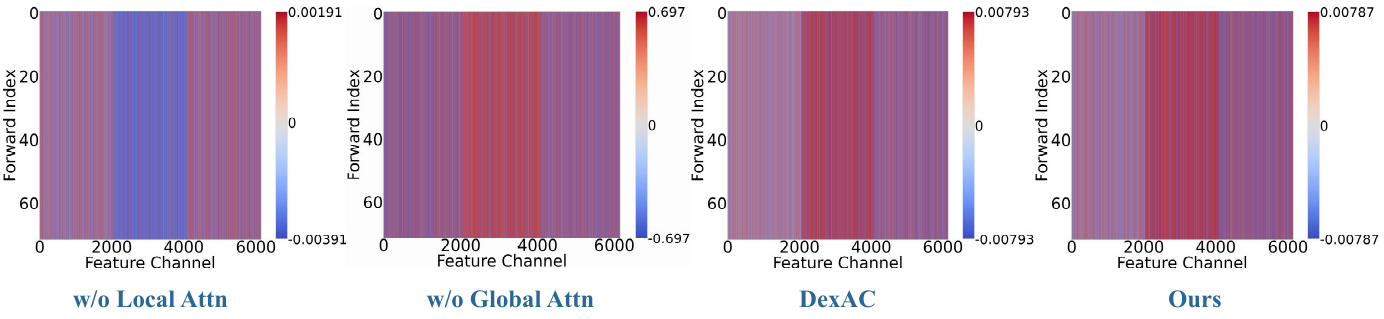}
    \caption{Action Heatmap in the AdaLN embedding
space.} 
    \label{fig:heatmap}
\end{figure}

\textbf{b. Are subtle action dimensions are effectively utilized?} To verify whether subtle action dimensions are effectively utilized, we visualize the action heatmap in the AdaLN embedding space. Specifically, we extract the action-conditioned embedding before AdaLN modulation and compute channel-wise activation magnitudes, forming a feature heatmap. As shown in Figure~\ref{fig:heatmap}, the \textbf{x-axis} represents feature channels and the \textbf{y-axis} denotes forward index, with color indicating activation strength. We compare four settings: (1) without Local Attention, (2) without Global Attention, (3) DexAC, and (4) Ours (DexAC+DINOv3). The baseline shows highly imbalanced activations dominated by a few channels, indicating that subtle action signals are suppressed. Removing global modulation leads to scattered and unstable patterns. In contrast, DexAC produces a more balanced and structured distribution, suggesting better preservation of dimension-wise action information. With DINO features, activations become more concentrated and semantically aligned, indicating stronger action-visual coupling. These results show that \textbf{our structured action conditioning improves the utilization of subtle action dimensions}, while \textbf{semantic visual features further enhance this effect}, leading to more balanced and physically meaningful action injection.

\begin{table*}[t]
\centering
\small
\renewcommand{\arraystretch}{1.1}
\caption{Ablation study on structured action conditioning components on EgoDex.}
\label{tab:depth_ablation}
\begin{tabular}{l|ccccccc}
\toprule
Components
& PSNR $\uparrow$ & SSIM $\uparrow$ & LPIPS $\downarrow$ & FID $\downarrow$ & FVD $\downarrow$ & PCK@10 $\uparrow$ & PCK@20 $\uparrow$ \\
\midrule
MLP Action Embed 
& 25.02 & 0.80 & 0.25 & 114.51 & 352.19 & 31.07 & 58.33 \\

w/o Local Attn 
& 25.12 & 0.80 & 0.25 & 114.51 & 377.03 & 30.48 & 58.50 \\

w/o Global Attn
& 24.82 & 0.79 & 0.26 & 116.31 & 419.90 & 30.81 & 54.21 \\

Full DexAC
& \textbf{25.13} & \textbf{0.80} & \textbf{0.24} & \textbf{106.67} & \textbf{284.40} & \textbf{32.70} & \textbf{60.59} \\
\bottomrule
\end{tabular}
\end{table*}

\begin{table}[H]
\centering
\small
\renewcommand{\arraystretch}{1.1}
\caption{Three action-family PCK evaluation.}
\label{tab:action_pck}
\begin{tabular}{l|cc|cc|cc}
\toprule
\multirow{2}{*}{Methods} & \multicolumn{2}{c|}{Wrist-Dom} & \multicolumn{2}{c|}{Finger-Dom} & \multicolumn{2}{c}{Head-Dom} \\
 & PCK@10 & PCK@20 & PCK@10 & PCK@20 & PCK@10 & PCK@20 \\
\midrule
MLP Action Embed  & 7.03  & 31.80 & 7.03  & 22.51 & 3.89  & 14.27 \\
w/o Global Attn & \textbf{88.83} & \textbf{93.76} & 0     & 0     & 0     & 0     \\
w/o Local Attn  & 48.88 & 72.60 & 29.03 & 53.87 & 35.75 & 56.03 \\
DexAC             & 33.04 & 58.42 & \textbf{33.04} & \textbf{58.42} & \underline{40.00} & \underline{60.99} \\
Ours (DexAC + DINOv3)    & \underline{54.65} & \underline{76.09} & \underline{29.68} & \underline{54.44} & \textbf{41.21} & \textbf{65.51} \\
\bottomrule
\end{tabular}
\end{table}

\begin{table*}[ht]
\centering
\small
\renewcommand{\arraystretch}{1.1}
\caption{Ablation study on action normalization and scaling on EgoDex.}
\label{tab:Norm_ablation}
\begin{tabular}{l|ccccccc}
\toprule
Methods
& PSNR $\uparrow$ & SSIM $\uparrow$ & LPIPS $\downarrow$ & FID $\downarrow$ & FVD $\downarrow$ & PCK@10 $\uparrow$ & PCK@20 $\uparrow$ \\
\midrule
Ours (w/o Norm)
& 24.69 & 0.79 & 0.26 & \textbf{117.54} & 418.53 & 31.19 & \textbf{56.64} \\

Ours (Scale 200)
& \textbf{24.79} & \textbf{0.80} & \textbf{0.25} & 118.22 & \textbf{371.18} & \textbf{31.42} & 56.40 \\
\bottomrule
\end{tabular}
\end{table*}
\begin{figure}[H]
    \centering
    \includegraphics[width=\linewidth]{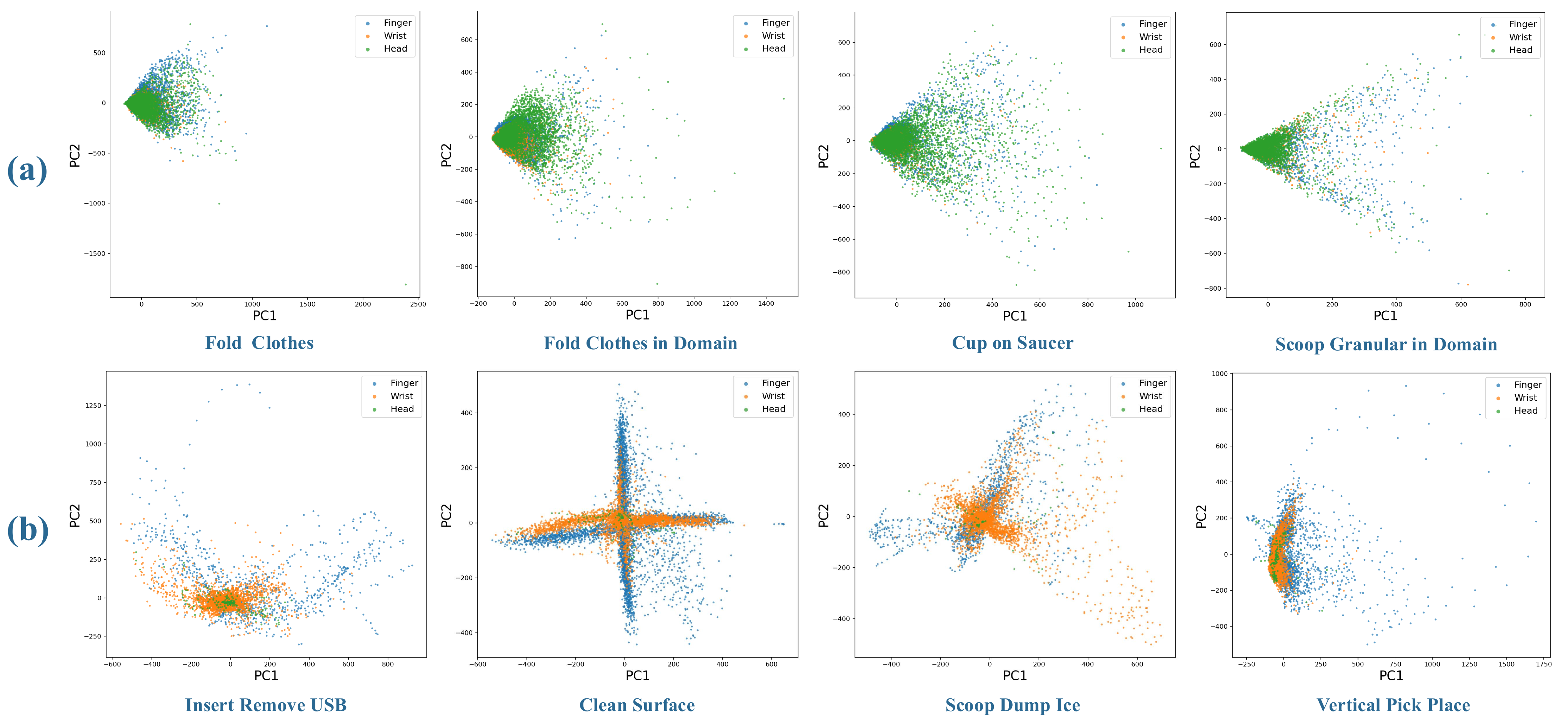}
    \caption{The PCA results of (a) EgoVerse and (b) EgoDex tasks.}   
    \label{fig:pca}
\end{figure}
\begin{figure}[H]
    \centering
    \includegraphics[width=\linewidth]{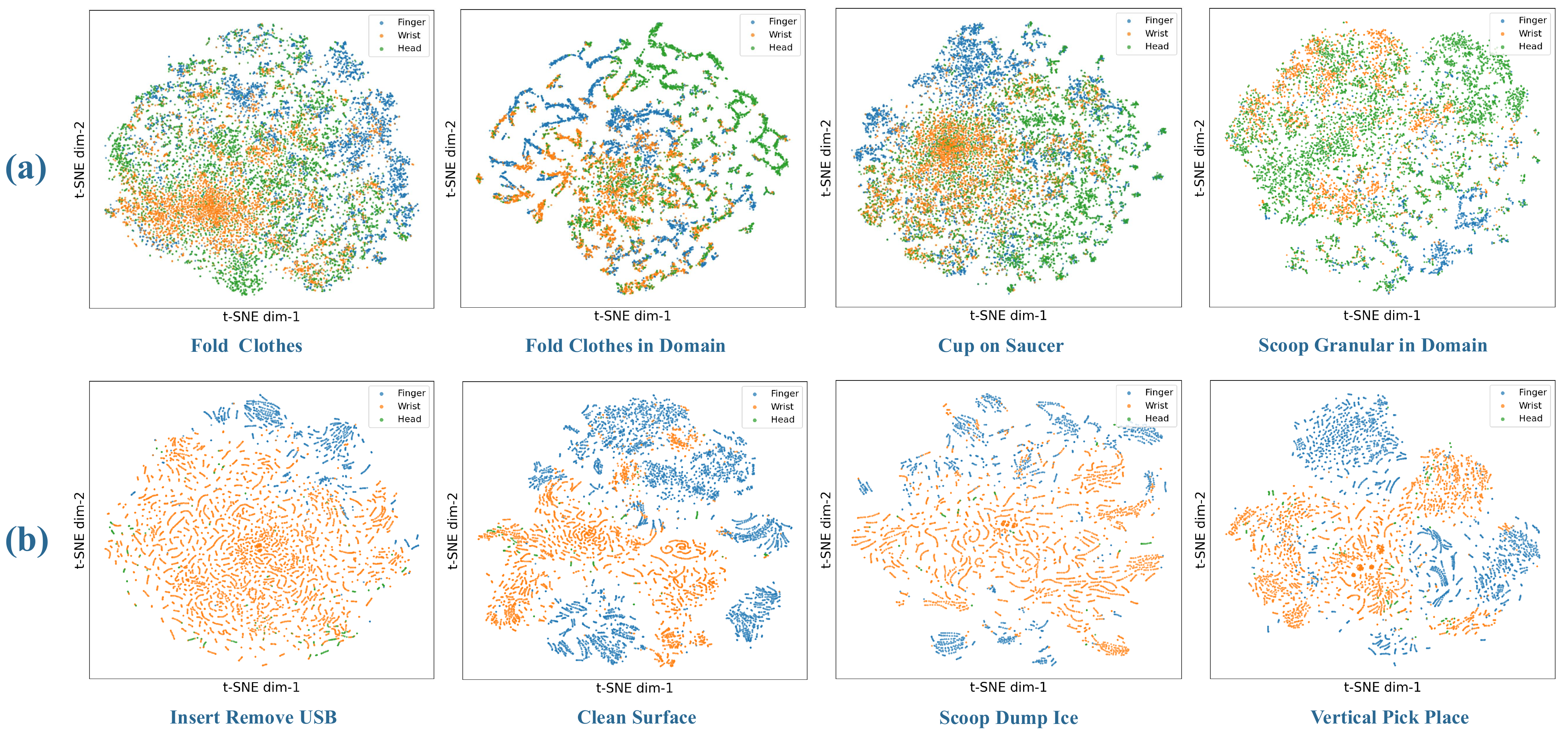}
    \caption{The t-SNE results of (a) EgoVerse and (b) EgoDex tasks.}
    \label{fig:tsne}
\end{figure}
\textbf{c. How do different action groups benefit?}
We provide action-family diagnostics across wrist-, finger-, and head-dominant groups to evaluate how different motions benefit from DexAC. Dimension-wise tokenization preserves subtle finger signals, while learnable global queries aggregate correlated wrist, finger, and camera motions. As shown in Table~\ref{tab:action_pck}, MLP action embedding remains weak across all three action-dominant cases, showing that flat global action injection limits action-following ability under heterogeneous high-DoF actions. In contrast, local attention refinement is biased toward specific motion groups, while DexAC+DINOv3 achieves the best average PCK across the three groups, demonstrating family-balanced physical consistency from structured action conditioning and semantic grounding. Table~\ref{tab:Norm_ablation} further verifies the importance of proper action normalization. Removing normalization degrades overall performance, while simply scaling all actions by 200 leads to a higher FVD than ours. This indicates that simple magnitude amplification can not effectively improve temporally consistent prediction for high-DoF dexterous actions. More importantly, per-dimension normalization is necessary but not sufficient. Although it alleviates raw numerical scale imbalance, it cannot prevent action semantics from collapsing under flat aggregation. Therefore, both scale alignment and structured action tokenization are essential for reliable high-DoF action conditioning.

% We provide action-family diagnostic across three action groups to evaluate how they benefit. Dimension-wise tokenization preserves subtle finger signals while learnable global queries still aggregate correlated wrist/finger/head motions. Table~\ref{tab:action_pck} shows that MLP embedding is weak across families, local-only refinement is biased, and DexAC+DINOv3 achieves the best average PCK across three action-dominant cases, demonstrating family-balanced physical consistency from structured action conditioning and semantic grounding. Table~\ref{tab:Norm_ablation} further verifies the importance of proper action normalization. Removing normalization degrades overall performance, while simply scaling actions by 200 (refer to Cosmos2.5-DROID) leads to a higher FVD than ours. This indicates that simple magnitude amplification does not help produce temporally consistent predictions for high-DoF dexterous actions.

\noindent We further visualize the PCA and t-SNE embeddings of the structured action tokenizer (SAT) features to examine how tokenizer represent high-DoF actions adaptively. The PCA results in Figure~\ref{fig:pca} reveal different distribution patterns across action groups. For example, finger-related samples exhibit both compact clusters and locally scattered points, suggesting that the tokenizer captures common finger-motion structures while remaining sensitive to task-dependent hand dynamics. The t-SNE visualization in Figure~\ref{fig:tsne} further shows that action features from different tasks form distinct but partially clustered distributions, indicating that the proposed tokenizer can separate both local and global action patterns across different motion groups. These observations support the effectiveness of dimension-wise structured representation for modeling heterogeneous high-DoF actions.

% Why Semantic Condition Help?
\textbf{d. Why does the semantic condition help?}
To analyze the effectiveness of semantic condition, we introduce two additional perceptual metrics and compare two variants in different weights of DINOv3 feature injection. Our results show that semantic conditioning improves perceptual fidelity, while changing the injection strength does not cause substantial performance fluctuations, although stronger semantic injection slightly degrades temporal consistency. This suggests that the performance variation in high-DoF dexterous prediction is more related to the lack of semantic grounding than to the exact injection strength. Meanwhile, combining DexAC further enhances DexAC-WM’s ability to perceive action semantics, improving visual perception quality while maintaining a lead in FVD and PCK. Results can refer to Table~\ref{tab:dreamsim_latentl2} and~\ref{tab:dino_scale} in Sec.~\ref{sec:semantic_cond}.

%We first vary DINOv3 injection weights to demonstrate the necessity of semantic condition, then introduce two additional perceptual metrics to evaluate performance gains of the semantic condition.

%证明DexAC-WM是否由于缺乏semantic而导致consistency的下降还是收到本身semantic注入强度的影响。

%Long-horizon Evaluation.
\textbf{e. Long horizon evaluation}. Then, we study 65 frames ($>$2s) for long-horizon evaluation. Table~\ref{tab:long_horizon} shows that FVD and PCK degrade substantially under longer rollouts. 

\begin{table*}[ht]
\centering
\small
\renewcommand{\arraystretch}{1.1}
\caption{Long horizon evaluation with 65 frames on EgoDex.}
\label{tab:long_horizon}
\begin{tabular}{l|ccccccc}
\toprule
Methods
& PSNR $\uparrow$ & SSIM $\uparrow$ & LPIPS $\downarrow$ & FID $\downarrow$ & FVD $\downarrow$ & PCK@10 $\uparrow$ & PCK@20 $\uparrow$ \\
\midrule
Cosmos-Predict2.5-2B (Base)
& 21.89 & 0.74 & 0.34 & 158.70 & 524.96 & 24.45 & 47.93 \\

Base + DINOv3
& \textbf{22.75} & \textbf{0.76} & \textbf{0.31} & \underline{134.12} & 499.44 & 26.45 & 41.73 \\

Base + DexAC
& \underline{22.48} & \underline{0.75} & \underline{0.32} & 134.28 & \underline{484.34} & \underline{27.22} & \textbf{50.70} \\

Ours (Base + DINOv3 + DexAC)
& 22.69 & \textbf{0.76} & \textbf{0.31} & \textbf{133.02} & \textbf{465.31} & \textbf{27.42} & \underline{50.67} \\
\bottomrule
\end{tabular}
\end{table*}

\section{Conclusion}
We propose DexAC-WM, a structured action-conditioning method that preserves dimension-level semantics and re-couples heterogeneous actions through local refinement and global modulation tailored to high-DoF dexterous world models. We further introduce additional semantic condition grounding and consistently improves DexAC-WM in visual-temporal quality and action consistency. Extensive experiments on both EgoDex and EgoVerse validate the effectiveness of combining structured action representation with semantic condition for high-DoF dexterous world modeling and the scalability of DexAC.

\vspace{1em}
\bibliography{ours-ref}

@String{Chelsea = "Chelsea" }

@article{yang2023learning,
  title={Learning interactive real-world simulators},
  author={Yang, Mengjiao and Du, Yilun and Ghasemipour, Kamyar and Tompson, Jonathan and Schuurmans, Dale and Abbeel, Pieter},
  journal={arXiv preprint arXiv:2310.06114},
  year={2023}
}

@inproceedings{bruce2024genie,
  title={Genie: Generative interactive environments},
  author={Bruce, Jake and Dennis, Michael D and Edwards, Ashley and Parker-Holder, Jack and Shi, Yuge and Hughes, Edward and Lai, Matthew and Mavalankar, Aditi and Steigerwald, Richie and Apps, Chris and others},
  booktitle={ICML},
  year={2024}
}

@article{zhou2024robodreamer,
  title={Robodreamer: Learning compositional world models for robot imagination},
  author={Zhou, Siyuan and Du, Yilun and Chen, Jiaben and Li, Yandong and Yeung, Dit-Yan and Gan, Chuang},
  journal={arXiv preprint arXiv:2404.12377},
  year={2024}
}

@article{zhu2025unified,
  title={Unified world models: Coupling video and action diffusion for pretraining on large robotic datasets},
  author={Zhu, Chuning and Yu, Raymond and Feng, Siyuan and Burchfiel, Benjamin and Shah, Paarth and Gupta, Abhishek},
  journal={arXiv preprint arXiv:2504.02792},
  year={2025}
}

@inproceedings{zhu2025irasim,
  title={Irasim: A fine-grained world model for robot manipulation},
  author={Zhu, Fangqi and Wu, Hongtao and Guo, Song and Liu, Yuxiao and Cheang, Chilam and Kong, Tao},
  booktitle={ICCV},
  year={2025}
}

@article{wang2026eva,
  title={EVA: Aligning Video World Models with Executable Robot Actions via Inverse Dynamics Rewards},
  author={Wang, Ruixiang and Liu, Qingming and Deng, Yueci and Liu, Guiliang and Liu, Zhen and Jia, Kui},
  journal={arXiv preprint arXiv:2603.17808},
  year={2026}
}

@article{zheng2025flare,
  title={Flare: Robot learning with implicit world modeling},
  author={Zheng, Ruijie and Wang, Jing and Reed, Scott and Bjorck, Johan and Fang, Yu and Hu, Fengyuan and Jang, Joel and Kundalia, Kaushil and Lin, Zongyu and Magne, Loic and others},
  journal={arXiv preprint arXiv:2505.15659},
  year={2025}
}

@inproceedings{nematollahi2025lumos,
  title={Lumos: Language-conditioned imitation learning with world models},
  author={Nematollahi, Iman and DeMoss, Branton and Chandra, Akshay L and Hawes, Nick and Burgard, Wolfram and Posner, Ingmar},
  booktitle={ICRA},
  year={2025}
}

@article{escoriza2025multi,
  title={Multi-Stage Manipulation with Demonstration-Augmented Reward, Policy, and World Model Learning},
  author={Escoriza, Adri{\`a} L{\'o}pez and Hansen, Nicklas and Tao, Stone and Mu, Tongzhou and Su, Hao},
  journal={arXiv preprint arXiv:2503.01837},
  year={2025}
}

@article{ning2025prompting,
  title={Prompting with the future: Open-world model predictive control with interactive digital twins},
  author={Ning, Chuanruo and Fang, Kuan and Ma, Wei-Chiu},
  journal={arXiv preprint arXiv:2506.13761},
  year={2025}
}

@article{kim2025dexterous,
  title={Dexterous World Models},
  author={Kim, Byungjun and Kim, Taeksoo and Lee, Junyoung and Joo, Hanbyul},
  journal={arXiv preprint arXiv:2512.17907},
  year={2025}
}

@misc{goswami2026worldmodelslearningdexterous,
      title={World Models for Learning Dexterous Hand-Object Interactions from Human Videos}, 
      author={Raktim Gautam Goswami and Amir Bar and David Fan and Tsung-Yen Yang and Gaoyue Zhou and Prashanth Krishnamurthy and Michael Rabbat and Farshad Khorrami and Yann LeCun},
      journal={arXiv preprint arXiv:2512.13644},
      year={2026}
}

@article{bai2025whole,
  title={Whole-body conditioned egocentric video prediction},
  author={Bai, Yutong and Tran, Danny and Bar, Amir and LeCun, Yann and Darrell, Trevor and Malik, Jitendra},
  journal={arXiv preprint arXiv:2506.21552},
  year={2025}
}

@article{sobinov2021neural,
  title={The neural mechanisms of manual dexterity},
  author={Sobinov, Anton R and Bensmaia, Sliman J},
  journal={Nature Reviews Neuroscience},
  year={2021}
}

@inproceedings{grauman2022ego4d,
  title={Ego4d: Around the world in 3,000 hours of egocentric video},
  author={Grauman, Kristen and Westbury, Andrew and Byrne, Eugene and Chavis, Zachary and Furnari, Antonino and Girdhar, Rohit and Hamburger, Jackson and Jiang, Hao and Liu, Miao and Liu, Xingyu and others},
  booktitle={CVPR},
  year={2022}
}

@inproceedings{damen2018scaling,
  title={Scaling egocentric vision: The epic-kitchens dataset},
  author={Damen, Dima and Doughty, Hazel and Farinella, Giovanni Maria and Fidler, Sanja and Furnari, Antonino and Kazakos, Evangelos and Moltisanti, Davide and Munro, Jonathan and Perrett, Toby and Price, Will and others},
  booktitle={ECCV},
  year={2018}
}

@inproceedings{liu2022hoi4d,
  title={Hoi4d: A 4d egocentric dataset for category-level human-object interaction},
  author={Liu, Yunze and Liu, Yun and Jiang, Che and Lyu, Kangbo and Wan, Weikang and Shen, Hao and Liang, Boqiang and Fu, Zhoujie and Wang, He and Yi, Li},
  booktitle={CVPR},
  year={2022}
}

@inproceedings{shaw2023videodex,
  title={Videodex: Learning dexterity from internet videos},
  author={Shaw, Kenneth and Bahl, Shikhar and Pathak, Deepak},
  booktitle={CoRL},
  year={2023}
}

@inproceedings{yu2025egosim,
  title={Egosim: Egocentric exploration in virtual worlds with multi-modal conditioning},
  author={Yu, Wei and Yin, Songheng and Easterbrook, Steve and Garg, Animesh},
  booktitle={ICLR},
  year={2025}
}

@inproceedings{grauman2024ego,
  title={Ego-exo4d: Understanding skilled human activity from first-and third-person perspectives},
  author={Grauman, Kristen and Westbury, Andrew and Torresani, Lorenzo and Kitani, Kris and Malik, Jitendra and Afouras, Triantafyllos and Ashutosh, Kumar and Baiyya, Vijay and Bansal, Siddhant and Boote, Bikram and others},
  booktitle={CVPR},
  year={2024}
}

@inproceedings{wu2023daydreamer,
  title={Daydreamer: World models for physical robot learning},
  author={Wu, Philipp and Escontrela, Alejandro and Hafner, Danijar and Abbeel, Pieter and Goldberg, Ken},
  booktitle={CoRL},
  year={2023}
}

@article{hu2023gaia,
  title={Gaia-1: A generative world model for autonomous driving},
  author={Hu, Anthony and Russell, Lloyd and Yeo, Hudson and Murez, Zak and Fedoseev, George and Kendall, Alex and Shotton, Jamie and Corrado, Gianluca},
  journal={arXiv preprint arXiv:2309.17080},
  year={2023}
}

@article{wang2024dexcap,
  title={Dexcap: Scalable and portable mocap data collection system for dexterous manipulation},
  author={Wang, Chen and Shi, Haochen and Wang, Weizhuo and Zhang, Ruohan and Fei-Fei, Li and Liu, C Karen},
  journal={arXiv preprint arXiv:2403.07788},
  year={2024}
}

@article{qin2023anyteleop,
  title={Anyteleop: A general vision-based dexterous robot arm-hand teleoperation system},
  author={Qin, Yuzhe and Yang, Wei and Huang, Binghao and Van Wyk, Karl and Su, Hao and Wang, Xiaolong and Chao, Yu-Wei and Fox, Dieter},
  journal={arXiv preprint arXiv:2307.04577},
  year={2023}
}

@inproceedings{qin2022dexmv,
  title={Dexmv: Imitation learning for dexterous manipulation from human videos},
  author={Qin, Yuzhe and Wu, Yueh-Hua and Liu, Shaowei and Jiang, Hanwen and Yang, Ruihan and Fu, Yang and Wang, Xiaolong},
  booktitle={ECCV},
  year={2022}
}

@inproceedings{qin2023dexpoint,
  title={Dexpoint: Generalizable point cloud reinforcement learning for sim-to-real dexterous manipulation},
  author={Qin, Yuzhe and Huang, Binghao and Yin, Zhao-Heng and Su, Hao and Wang, Xiaolong},
  booktitle={CoRL},
  year={2023}
}

@inproceedings{handa2023dextreme,
  title={Dextreme: Transfer of agile in-hand manipulation from simulation to reality},
  author={Handa, Ankur and Allshire, Arthur and Makoviychuk, Viktor and Petrenko, Aleksei and Singh, Ritvik and Liu, Jingzhou and Makoviichuk, Denys and Van Wyk, Karl and Zhurkevich, Alexander and Sundaralingam, Balakumar and others},
  booktitle={ICRA},
  year={2023}
}

@article{wen2023any,
  title={Any-point trajectory modeling for policy learning},
  author={Wen, Chuan and Lin, Xingyu and So, John and Chen, Kai and Dou, Qi and Gao, Yang and Abbeel, Pieter},
  journal={arXiv preprint arXiv:2401.00025},
  year={2023}
}

@inproceedings{ye2023affordance,
  title={Affordance diffusion: Synthesizing hand-object interactions},
  author={Ye, Yufei and Li, Xueting and Gupta, Abhinav and De Mello, Shalini and Birchfield, Stan and Song, Jiaming and Tulsiani, Shubham and Liu, Sifei},
  booktitle={CVPR},
  year={2023}
}

@article{Ma_He_Cun_Wang_Chen_Li_Chen_2024,  title={Follow Your Pose: Pose-Guided Text-to-Video Generation Using Pose-Free Videos},  
    author={Ma, Yue and He, Yingqing and Cun, Xiaodong and Wang, Xintao and Chen, Siran and Li, Xiu and Chen, Qifeng}, 
    journal={AAAI},
    year={2024}
   }

@article{hansen2023td,
  title={Td-mpc2: Scalable, robust world models for continuous control},
  author={Hansen, Nicklas and Su, Hao and Wang, Xiaolong},
  journal={arXiv preprint arXiv:2310.16828},
  year={2023}
}

@article{guo2025ctrl,
  title={Ctrl-world: A controllable generative world model for robot manipulation},
  author={Guo, Yanjiang and Shi, Lucy Xiaoyang and Chen, Jianyu and Finn, Chelsea},
  journal={arXiv preprint arXiv:2510.10125},
  year={2025}
}

@article{zheng2025survey,
  title={A survey of embodied learning for object-centric robotic manipulation},
  author={Zheng, Ying and Yao, Lei and Su, Yuejiao and Zhang, Yi and Wang, Yi and Zhao, Sicheng and Zhang, Yiyi and Chau, Lap-Pui},
  journal={Machine Intelligence Research},
  year={2025}
}

@article{an2025dexterous,
  title={Dexterous manipulation through imitation learning: A survey},
  author={An, Shan and Meng, Ziyu and Tang, Chao and Zhou, Yuning and Liu, Tengyu and Ding, Fangqiang and Zhang, Shufang and Mu, Yao and Song, Ran and Zhang, Wei and others},
  journal={arXiv preprint arXiv:2504.03515},
  year={2025}
}

@article{li2022survey,
  title={A survey of multifingered robotic manipulation: Biological results, structural evolvements, and learning methods},
  author={Li, Yinlin and Wang, Peng and Li, Rui and Tao, Mo and Liu, Zhiyong and Qiao, Hong},
  journal={Frontiers in Neurorobotics},
  year={2022}
}

@article{simeoni2025dinov3,
  title={Dinov3},
  author={Sim{\'e}oni, Oriane and Vo, Huy V and Seitzer, Maximilian and Baldassarre, Federico and Oquab, Maxime and Jose, Cijo and Khalidov, Vasil and Szafraniec, Marc and Yi, Seungeun and Ramamonjisoa, Micha{\"e}l and others},
  journal={arXiv preprint arXiv:2508.10104},
  year={2025}
}

@article{ali2025world,
  title={World simulation with video foundation models for physical ai},
  author={Ali, Arslan and Bai, Junjie and Bala, Maciej and Balaji, Yogesh and Blakeman, Aaron and Cai, Tiffany and Cao, Jiaxin and Cao, Tianshi and Cha, Elizabeth and Chao, Yu-Wei and others},
  journal={arXiv preprint arXiv:2511.00062},
  year={2025}
}

@article{azzolini2025cosmos,
  title={Cosmos-reason1: From physical common sense to embodied reasoning},
  author={Azzolini, Alisson and Bai, Junjie and Brandon, Hannah and Cao, Jiaxin and Chattopadhyay, Prithvijit and Chen, Huayu and Chu, Jinju and Cui, Yin and Diamond, Jenna and Ding, Yifan and others},
  journal={arXiv preprint arXiv:2503.15558},
  year={2025}
}

@article{wan2025wan,
  title={Wan: Open and advanced large-scale video generative models},
  author={Wan, Team and Wang, Ang and Ai, Baole and Wen, Bin and Mao, Chaojie and Xie, Chen-Wei and Chen, Di and Yu, Feiwu and Zhao, Haiming and Yang, Jianxiao and others},
  journal={arXiv preprint arXiv:2503.20314},
  year={2025}
}

@article{hoque2025egodex,
  title={Egodex: Learning dexterous manipulation from large-scale egocentric video},
  author={Hoque, Ryan and Huang, Peide and Yoon, David J and Sivapurapu, Mouli and Zhang, Jian},
  journal={arXiv preprint arXiv:2505.11709},
  year={2025}
}

@inproceedings{bar2025navigation,
  title={Navigation world models},
  author={Bar, Amir and Zhou, Gaoyue and Tran, Danny and Darrell, Trevor and LeCun, Yann},
  booktitle={CVPR},
  year={2025}
}

@inproceedings{hore2010image,
  title={Image quality metrics: PSNR vs. SSIM},
  author={Hore, Alain and Ziou, Djemel},
  booktitle={ICPR},
  year={2010}
}

@inproceedings{zhang2018unreasonable,
  title={The unreasonable effectiveness of deep features as a perceptual metric},
  author={Zhang, Richard and Isola, Phillip and Efros, Alexei A and Shechtman, Eli and Wang, Oliver},
  booktitle={CVPR},
  year={2018}
}

@article{wang2004image,
  title={Image quality assessment: from error visibility to structural similarity},
  author={Wang, Zhou and Bovik, Alan C and Sheikh, Hamid R and Simoncelli, Eero P},
  journal={IEEE TIP},
  year={2004}
}

@article{heusel2017gans,
  title={Gans trained by a two time-scale update rule converge to a local nash equilibrium},
  author={Heusel, Martin and Ramsauer, Hubert and Unterthiner, Thomas and Nessler, Bernhard and Hochreiter, Sepp},
  journal={NeurIPS},
  year={2017}
}

@article{unterthiner2018towards,
  title={Towards accurate generative models of video: A new metric \& challenges},
  author={Unterthiner, Thomas and Van Steenkiste, Sjoerd and Kurach, Karol and Marinier, Raphael and Michalski, Marcin and Gelly, Sylvain},
  journal={arXiv preprint arXiv:1812.01717},
  year={2018}
}

@article{yang2012articulated,
  title={Articulated human detection with flexible mixtures of parts},
  author={Yang, Yi and Ramanan, Deva},
  journal={IEEE TPAMI},
  year={2012}
}

@article{assran2025v,
  title={V-jepa 2: Self-supervised video models enable understanding, prediction and planning},
  author={Assran, Mido and Bardes, Adrien and Fan, David and Garrido, Quentin and Howes, Russell and Muckley, Matthew and Rizvi, Ammar and Roberts, Claire and Sinha, Koustuv and Zholus, Artem and others},
  journal={arXiv preprint arXiv:2506.09985},
  year={2025}
}

@article{hafner2023mastering,
  title={Mastering diverse domains through world models},
  author={Hafner, Danijar and Pasukonis, Jurgis and Ba, Jimmy and Lillicrap, Timothy},
  journal={arXiv preprint arXiv:2301.04104},
  year={2023}
}

@article{schrittwieser2020mastering,
  title={Mastering atari, go, chess and shogi by planning with a learned model},
  author={Schrittwieser, Julian and Antonoglou, Ioannis and Hubert, Thomas and Simonyan, Karen and Sifre, Laurent and Schmitt, Simon and Guez, Arthur and Lockhart, Edward and Hassabis, Demis and Graepel, Thore and others},
  journal={Nature},
  year={2020}
}

@inproceedings{peebles2023scalable,
  title={Scalable diffusion models with transformers},
  author={Peebles, William and Xie, Saining},
  booktitle={ICCV},
  year={2023}
}

@article{chen2021decision,
  title={Decision transformer: Reinforcement learning via sequence modeling},
  author={Chen, Lili and Lu, Kevin and Rajeswaran, Aravind and Lee, Kimin and Grover, Aditya and Laskin, Misha and Abbeel, Pieter and Srinivas, Aravind and Mordatch, Igor},
  journal={NeurIPS},
  year={2021}
}

@article{xiao2022masked,
  title={Masked visual pre-training for motor control},
  author={Xiao, Tete and Radosavovic, Ilija and Darrell, Trevor and Malik, Jitendra},
  journal={arXiv preprint arXiv:2203.06173},
  year={2022}
}

@article{nair2022r3m,
  title={R3m: A universal visual representation for robot manipulation},
  author={Nair, Suraj and Rajeswaran, Aravind and Kumar, Vikash and Finn, Chelsea and Gupta, Abhinav},
  journal={arXiv preprint arXiv:2203.12601},
  year={2022}
}

@article{ma2022vip,
  title={Vip: Towards universal visual reward and representation via value-implicit pre-training},
  author={Ma, Yecheng Jason and Sodhani, Shagun and Jayaraman, Dinesh and Bastani, Osbert and Kumar, Vikash and Zhang, Amy},
  journal={arXiv preprint arXiv:2210.00030},
  year={2022}
}

@article{zhou2024dino,
  title={Dino-wm: World models on pre-trained visual features enable zero-shot planning},
  author={Zhou, Gaoyue and Pan, Hengkai and LeCun, Yann and Pinto, Lerrel},
  journal={arXiv preprint arXiv:2411.04983},
  year={2024}
}

@inproceedings{zhu2023vima,
  title={Vima: General robot manipulation with multimodal prompts},
  author={Zhu, Y and others},
  booktitle={ICLR},
  year={2023}
}

@article{chen2025internvla,
  title={Internvla-m1: A spatially guided vision-language-action framework for generalist robot policy},
  author={Chen, Xinyi and Chen, Yilun and Fu, Yanwei and Gao, Ning and Jia, Jiaya and Jin, Weiyang and Li, Hao and Mu, Yao and Pang, Jiangmiao and Qiao, Yu and others},
  journal={arXiv preprint arXiv:2510.13778},
  year={2025}
}

@inproceedings{wang2025vggt,
  title={Vggt: Visual geometry grounded transformer},
  author={Wang, Jianyuan and Chen, Minghao and Karaev, Nikita and Vedaldi, Andrea and Rupprecht, Christian and Novotny, David},
  booktitle={CVPR},
  year={2025}
}

@article{fu2023dreamsim,
  title={Dreamsim: Learning new dimensions of human visual similarity using synthetic data},
  author={Fu, Stephanie and Tamir, Netanel and Sundaram, Shobhita and Chai, Lucy and Zhang, Richard and Dekel, Tali and Isola, Phillip},
  journal={arXiv preprint arXiv:2306.09344},
  year={2023}
}

@inproceedings{rombach2022high,
  title={High-resolution image synthesis with latent diffusion models},
  author={Rombach, Robin and Blattmann, Andreas and Lorenz, Dominik and Esser, Patrick and Ommer, Bj{\"o}rn},
  booktitle={CVPR},
  year={2022}
}

@article{punamiya2026egoverse,
  title={Egoverse: An egocentric human dataset for robot learning from around the world},
  author={Punamiya, Ryan and Kareer, Simar and Liu, Zeyi and Citron, Josh and Qiu, Ri-Zhao and Cai, Xiongyi and Gavryushin, Alexey and Chen, Jiaqi and Liconti, Davide and Zhu, Lawrence Y and others},
  journal={arXiv preprint arXiv:2604.07607},
  year={2026}
}

@inproceedings{banerjee2025hot3d,
  title={Hot3d: Hand and object tracking in 3d from egocentric multi-view videos},
  author={Banerjee, Prithviraj and Shkodrani, Sindi and Moulon, Pierre and Hampali, Shreyas and Han, Shangchen and Zhang, Fan and Zhang, Linguang and Fountain, Jade and Miller, Edward and Basol, Selen and others},
  booktitle={CVPR},
  year={2025}
}

@inproceedings{kareer2025egomimic,
  title={Egomimic: Scaling imitation learning via egocentric video},
  author={Kareer, Simar and Patel, Dhruv and Punamiya, Ryan and Mathur, Pranay and Cheng, Shuo and Wang, Chen and Hoffman, Judy and Xu, Danfei},
  booktitle={ICRA},
  year={2025},
}

@article{jawaid2025openego,
  title={Openego: A large-scale multimodal egocentric dataset for dexterous manipulation},
  author={Jawaid, Ahad and Xiang, Yu},
  journal={arXiv preprint arXiv:2509.05513},
  year={2025}
}

@article{zheng2026egoscale,
  title={Egoscale: Scaling dexterous manipulation with diverse egocentric human data},
  author={Zheng, Ruijie and Niu, Dantong and Xie, Yuqi and Wang, Jing and Xu, Mengda and Jiang, Yunfan and Casta{\~n}eda, Fernando and Hu, Fengyuan and Tan, You Liang and Fu, Letian and others},
  journal={arXiv preprint arXiv:2602.16710},
  year={2026}
}

@article{sun2026vggt,
  title={Vggt-world: Transforming vggt into an autoregressive geometry world model},
  author={Sun, Xiangyu and Wang, Shijie and Zhang, Fengyi and Liu, Lin and Jia, Caiyan and Song, Ziying and Huang, Zi and Luo, Yadan},
  journal={arXiv preprint arXiv:2603.12655},
  year={2026}
}

@article{zhou2025omniworld,
  title={Omniworld: A multi-domain and multi-modal dataset for 4d world modeling},
  author={Zhou, Yang and Wang, Yifan and Zhou, Jianjun and Chang, Wenzheng and Guo, Haoyu and Li, Zizun and Ma, Kaijing and Li, Xinyue and Wang, Yating and Zhu, Haoyi and others},
  journal={arXiv preprint arXiv:2509.12201},
  year={2025}
}

@article{agarwal2025cosmos,
  title={Cosmos world foundation model platform for physical ai},
  author={Agarwal, Niket and Ali, Arslan and Bala, Maciej and Balaji, Yogesh and Barker, Erik and Cai, Tiffany and Chattopadhyay, Prithvijit and Chen, Yongxin and Cui, Yin and Ding, Yifan and others},
  journal={arXiv preprint arXiv:2501.03575},
  year={2025}
}
\clearpage
\appendix

\section{ Overview of the Appendix}
\label{sec:overview}
\par This appendix contains additional analysis, experimental details, and discussions, organized as follows:

\noindent $\bullet$ Sec.~\ref{sec:appendix_details} outlines the additional implementation details in experiments.\\
\noindent $\bullet$ Sec.~\ref{sec:dexac} presents model performance and effects of DexAC.\\
\noindent $\bullet$ Sec.~\ref{sec:semantic_cond} investigates the effectiveness of the semantic condition branch.\\
\noindent $\bullet$ Sec.~\ref{sec:appendix_qualitative} visualizes more qualitative results for further comparison and analysis.\\
\noindent $\bullet$ Sec.~\ref{sec:limitation} discusses the limitations of the current study and highlights possible directions for future work.

\section{Additional Implementation Details}
\label{sec:appendix_details}
All models are trained under the Fully Sharded Data Parallel (FSDP) setting for 2 epochs (about 20 hours per epoch in Egodex). We report results using the final checkpoint for all methods. The remaining configuration is identical to that described in the main paper.

% \section{Dataset and Environment Setup}
% \label{sec:appendix_dataset}
% Details about the data filtering process and the simulator configurations...
% \section{Details of our backbone architecture}
% \label{sec:dit}
% We present the details of our backbone architecture in Figures~\ref{fig:dit}. 

\section{Model Performance and Effects of DexAC}
\label{sec:dexac}

\begin{figure}[h]
    \centering
    \includegraphics[width=\linewidth]{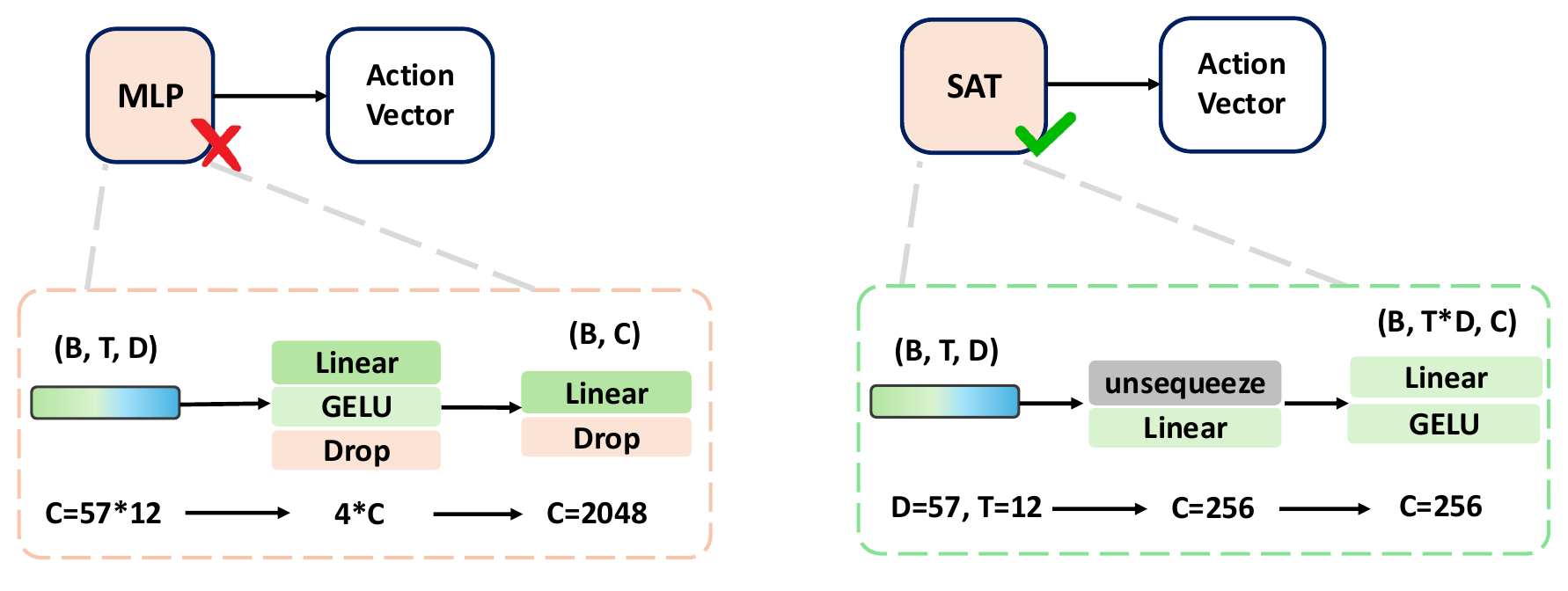}
    \vspace{-10pt}
    \caption{Details comparison of the vanilla MLP action embedding method and our proposed structured action tokenizer (SAT) for action dimension-level feature preservation.}    
    \label{fig:tokenizer}
\end{figure}

% Figure~\ref{fig:dit} provides additional details of the DiT backbone architecture, which consists of 28 blocks, and illustrates how the structured action condition is injected into each block through Adaptive Layer Normalization (AdaLN).
% \begin{wrapfigure}{l}{0.5\textwidth}
%   \vspace{-10pt}
%   \centering
%   \includegraphics[width=0.48\textwidth]{dit.pdf}
%   \caption{\textbf{Details of our backbone architecture.}}
%   \label{fig:dit}
%   \vspace{-10pt}
% \end{wrapfigure} 

we provide more details of the structured action tokenizer (SAT) of DexAC in Figure~\ref{fig:tokenizer}. We set the action embedding space of 256, with 12 action chunk-size T. We evaluate different variants of action of the parameters and tokenization and computation cost in Table~\ref{tab:action_param_comparison}. The results show that our proposed full DexAC achieve lower parameters than MLP-embedding baseline. And full DexAC requires 3.55G FLOPs
mainly from local refinement, but it is more parameter
efficient than the MLP baseline and yields clear gains, making the extra computation a reasonable cost. Furthermore, without global Attention or local attention shows significant drop of DreamSim \cite{fu2023dreamsim} and Latent L2 \cite{rombach2022high} which indicates the effectiveness of two components for action refinement and perceptual gains. This suggests that the gains of DexAC are not simply due to increased parameter count, but instead arise from a more effective structured action-conditioning design.

\begin{table}[h]
\centering

\renewcommand{\arraystretch}{1.1} 
\caption{Model complexity comparison: Parameters and FLOPs.}
\label{tab:action_param_comparison}
\begin{tabular}{l|cc}
\toprule
Components & Params & FLOPs  \\
\midrule
MLP Action Embed      & 22.39M & 24.64M  \\
w/ SAT    & 0.512K    & 245.76K \\
w/ Local Attn & 13.65M & 3.42G   \\
w/ Global Attn & 2.37M  & 126.45M \\
\rowcolor{rowgray}
Full DexAC & \textbf{16.02M} & \textbf{3.55G} \\
\bottomrule
\end{tabular}
\end{table}

\begin{table}[h]
\caption{Additional Quantitative Results of DreamSim and Latent L2. Lower is better for both metrics.}
\renewcommand{\arraystretch}{1.1}
\label{tab:dreamsim_latentl2}
\centering
\begin{tabular}{l|cc}
\toprule
Baselines & DreamSim $\downarrow$ & Latent L2 $\downarrow$ \\
\midrule
Cosmos-Predict2.5-2B (Base) & 0.1245 & 9.3637 \\
\midrule
Base + DINOv3 & 0.1233 & 9.2491 \\
Base + DexAC & 0.1235 & 9.2518 \\
\rowcolor{rowgray}
Base + DexAC + DINOv3 (Ours) & \textbf{0.1229} & \textbf{9.2304} \\
\midrule
w/o Local Attn & 0.1275 & 9.4113 \\
w/o Global Attn & 0.1271 & 9.3537\\
\bottomrule
\end{tabular}
\end{table}

\section{Effects of Semantic Condition}
\label{sec:semantic_cond}
We vary the DINOv3 injection strengths to compare a moderate conditioning strength against a substantially stronger setting to examine whether increasing semantic guidance consistently benefits perception quality. Table~\ref{tab:dino_scale} shows that increasing the weight of the semantic branch from 0.2 to 0.6 has little effect on the first four metrics, while FVD shows a slight increase. To further evaluate the performance gain of the semantic condition in perception quality, we evaluate DreamSim~\cite{fu2023dreamsim} and Latent L2~\cite{rombach2022high} in Table~\ref{tab:dreamsim_latentl2}. These findings indicate that DINOv3 provides useful action-semantic knowledge. When combined with DexAC, these priors are further aligned with structured, high-DoF action features, enabling DexAC-WM to learn a semantic-aware action representation that better preserves visual changes during fine-grained dexterous motions.

% We apply two weight settings (0.2 and 0.6) to compare a moderate semantic-conditioning strength against a substantially stronger setting, in order to examine whether increasing DINO guidance consistently benefits generation quality. Table~\ref{tab:dino_scale} shows that increasing the DINO branch scale from 0.2 to 0.6 brings slight improvements in PSNR and FID, but degrades LPIPS and FVD, indicating a trade-off between spatial fidelity and perceptual-temporal coherence. Therefore, we adopt 0.2 as the default setting for its more balanced overall performance. To further evaluate the performance of perceptual capability of DINO branch in the semantic condition, we additionally evaluate the DreamSim~\cite{fu2023dreamsim} and Latent L2~\cite{rombach2022high} in Table~\ref{tab:dreamsim_latentl2}. The results shows that the base model with semantic condition achieves consistent improvements compared with the base model, and the full model achieves the best results. This observation is consistent with our main findings: semantic conditioning mainly benefits visual perceptual quality, while DexAC further improves performance through a more effective structured action-conditioning design.

\begin{table}[h]
\caption{Ablation results of the weight of our semantic condition.}
\renewcommand{\arraystretch}{1.1}
\label{tab:dino_scale}
\centering
\begin{tabular}{l|ccccc}
\toprule
Weights & PSNR $\uparrow$ & SSIM $\uparrow$ & LPIPS $\downarrow$ & FID $\downarrow$ & FVD $\downarrow$ \\
\midrule
\rowcolor{rowgray}
0.2 (Ours) & 25.13 & \textbf{0.80} & \textbf{0.24} & 106.67 & \textbf{284.40} \\
0.6 & \textbf{25.14} & 0.80 & 0.25 & \textbf{106} & 288.17 \\
\bottomrule
\end{tabular}
\end{table}

\section{Additional Qualitative Results}
\label{sec:appendix_qualitative}
\begin{figure}[ht]
    \centering
    \includegraphics[width=\linewidth]{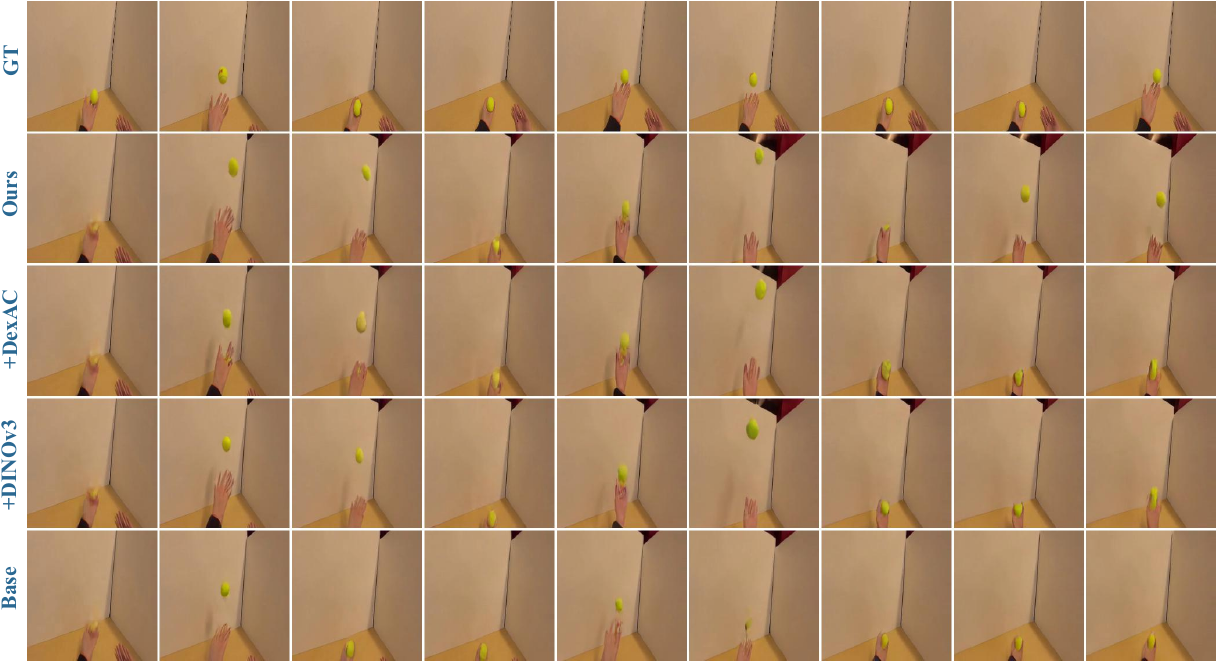}
    \vspace{-10pt}
    \caption{Throwing a Tennis Ball.}    
    \label{fig:failure_ball}
\end{figure}

\begin{figure*}[t]
    \centering\includegraphics[width=\textwidth]{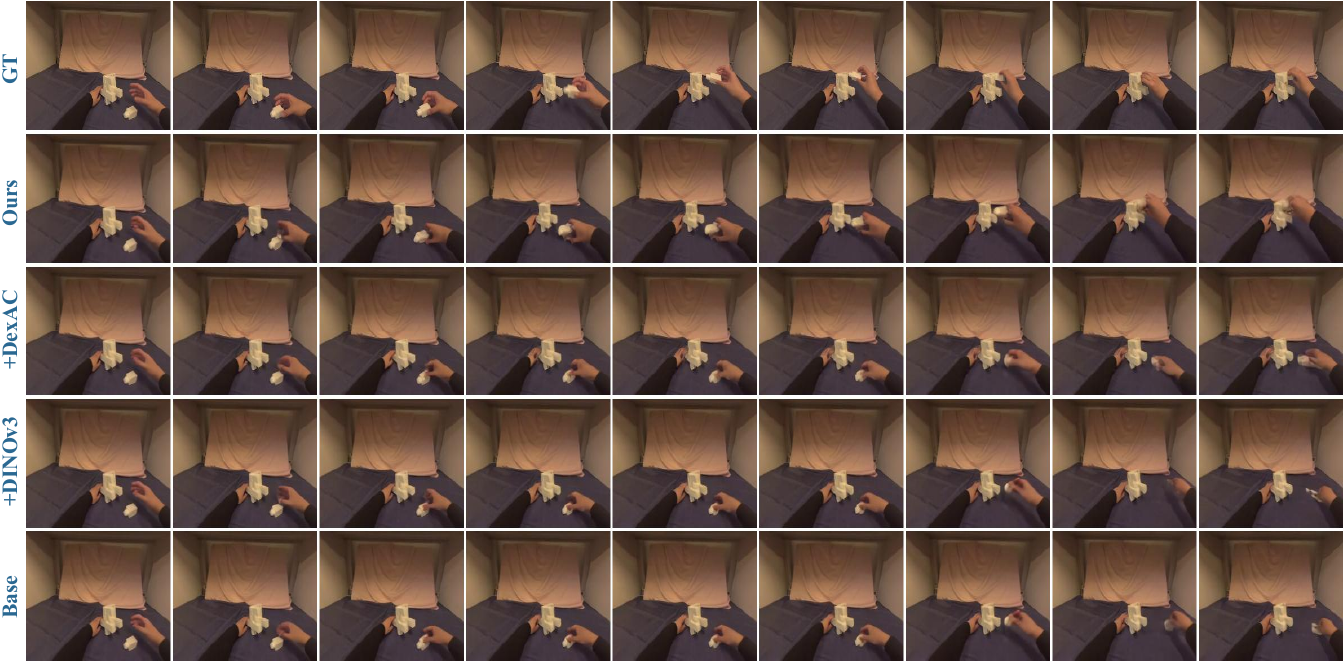}
    \vspace{-10pt}
    \caption{Assemble Disassemble Furniture Bench Chair.} \label{fig:26_1}
\end{figure*}

\begin{figure*}[t]
    \centering\includegraphics[width=\textwidth]{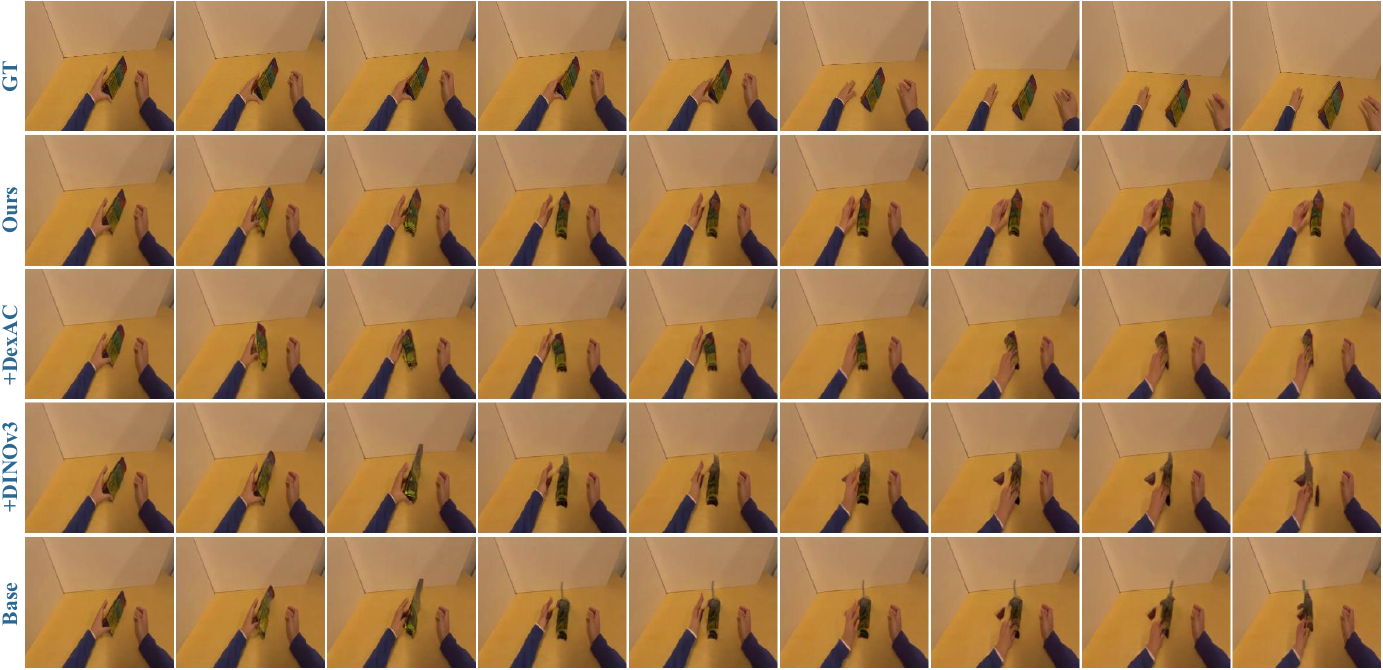}
    \vspace{-10pt}
    \caption{Assemble Disassemble Tiles.} \label{fig:26_2}
\end{figure*}

\begin{figure*}[t]
    \centering\includegraphics[width=\textwidth]{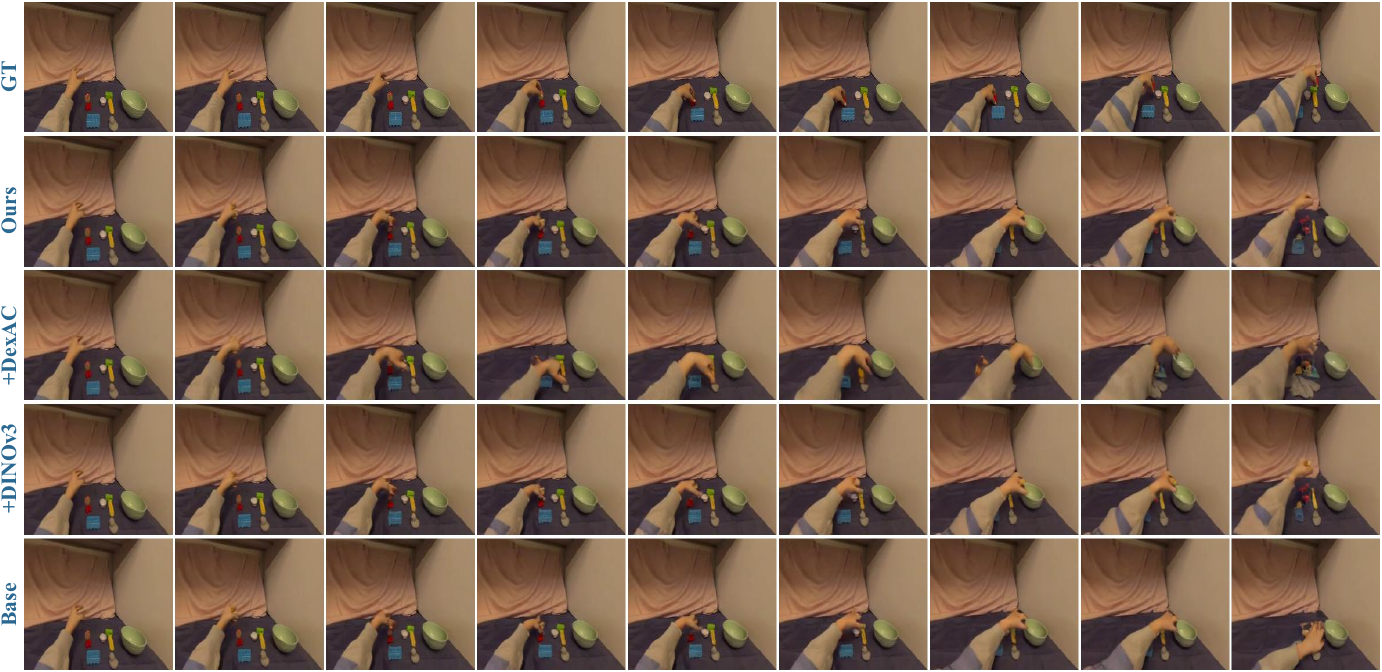}
    \vspace{-10pt}
    \caption{Basic Pick Place.} \label{fig:26_3}
\end{figure*}

\begin{figure*}[t]
    \centering\includegraphics[width=\textwidth]{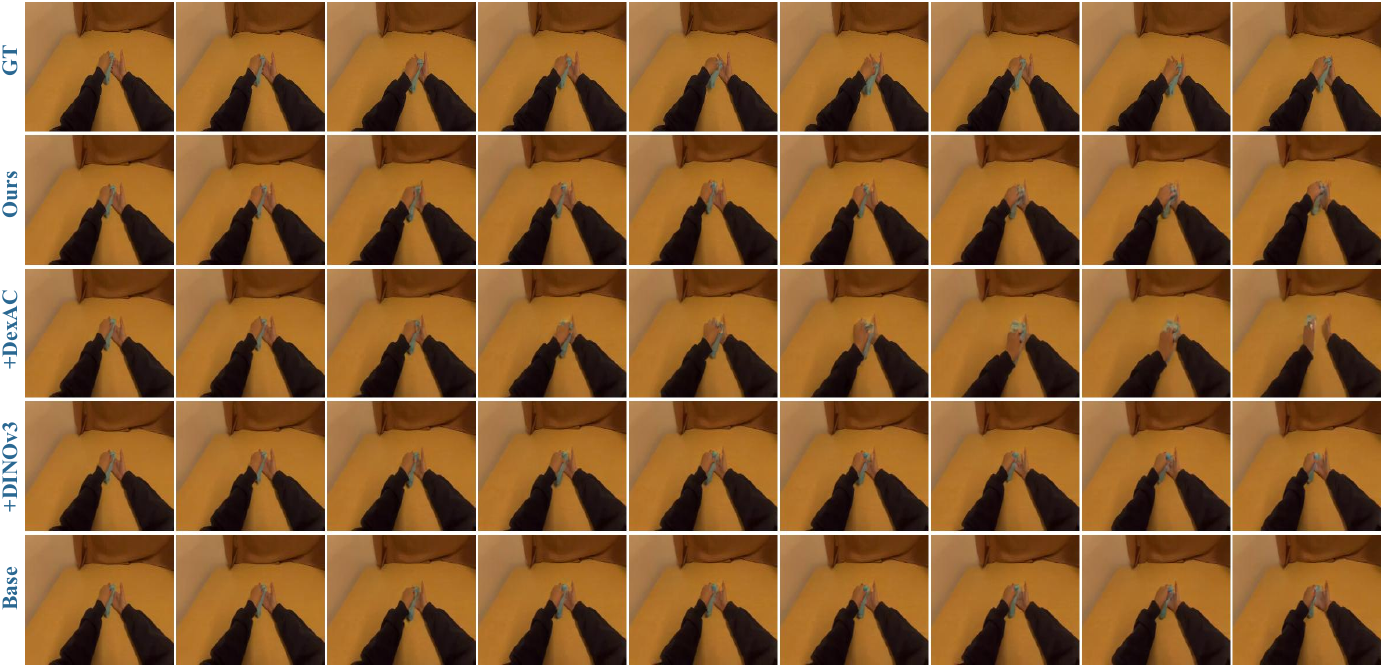}
    \vspace{-10pt}
    \caption{Dry Hands.} \label{fig:26_4}
\end{figure*}

\begin{figure*}[t]
    \centering\includegraphics[width=\textwidth]{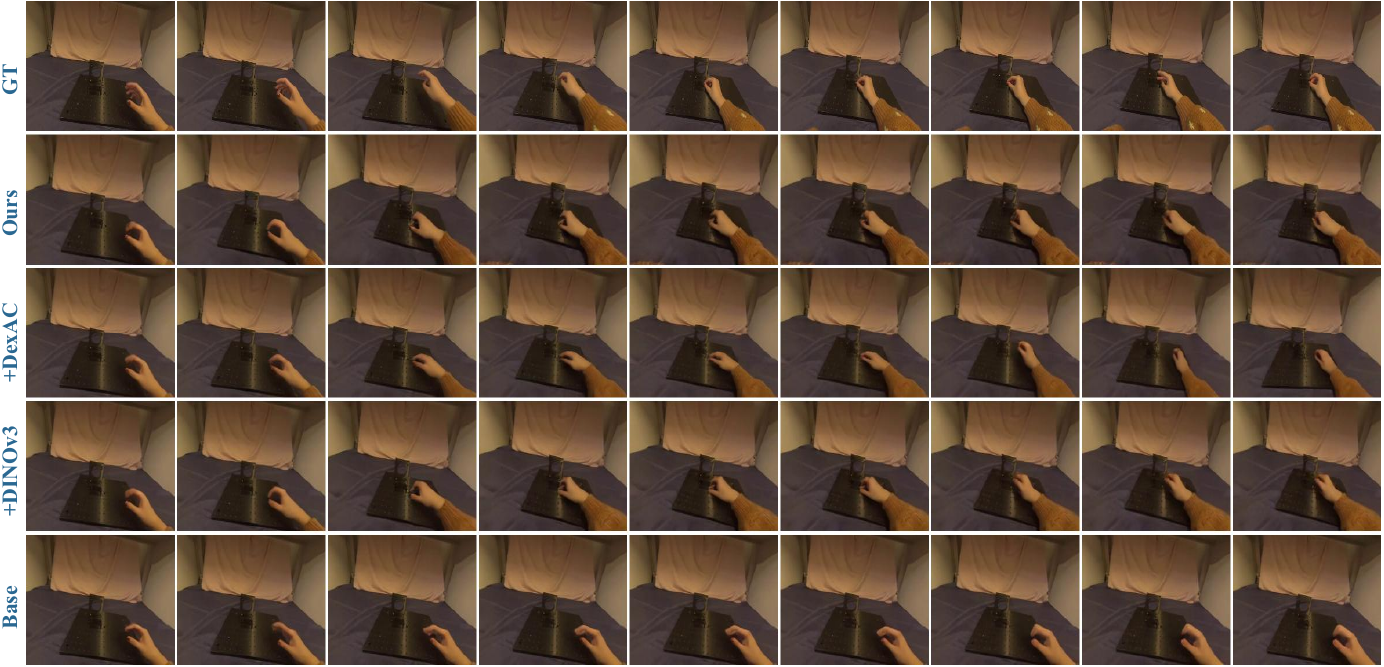}
    \vspace{-10pt}
    \caption{Screw Unscrew Fingers Fixture.} \label{fig:26_6}
\end{figure*}

\begin{figure*}[t]
    \centering\includegraphics[width=\textwidth]{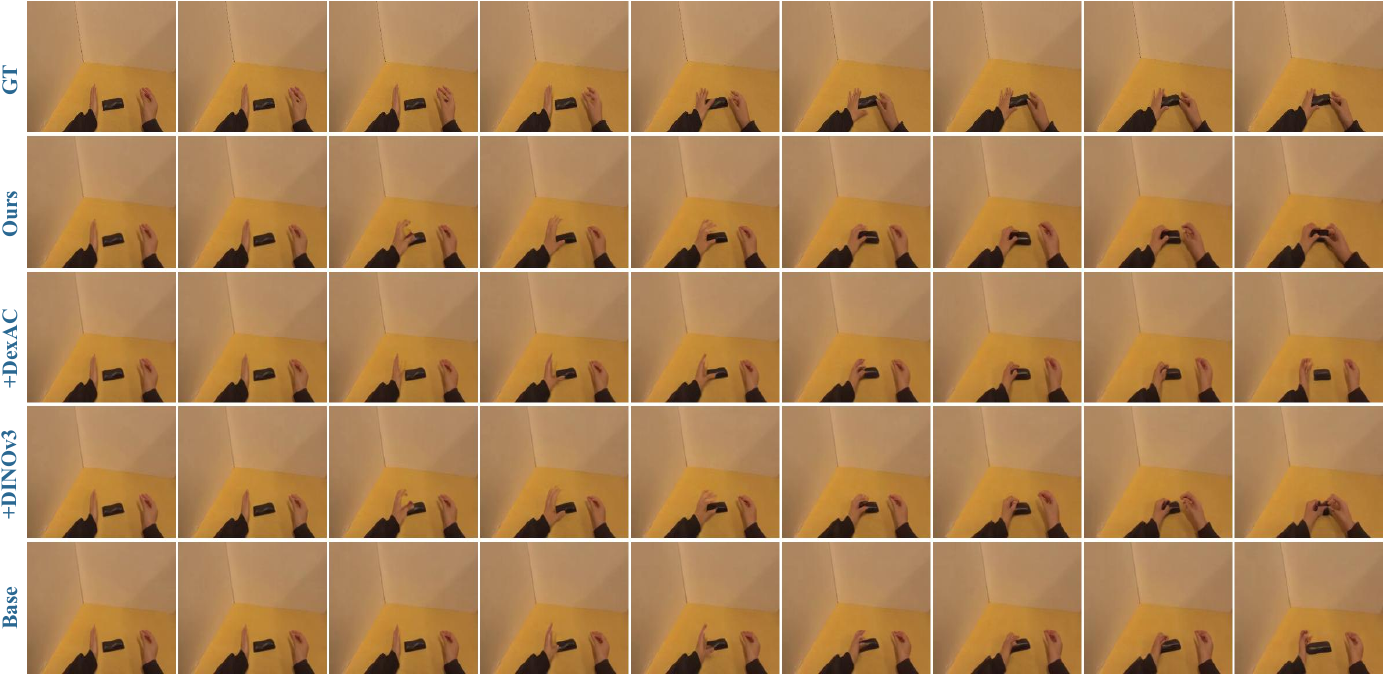}
    \vspace{-10pt}
    \caption{Slot Batteries.} \label{fig:26_7}
\end{figure*}

\begin{figure*}[t]
    \centering\includegraphics[width=\textwidth]{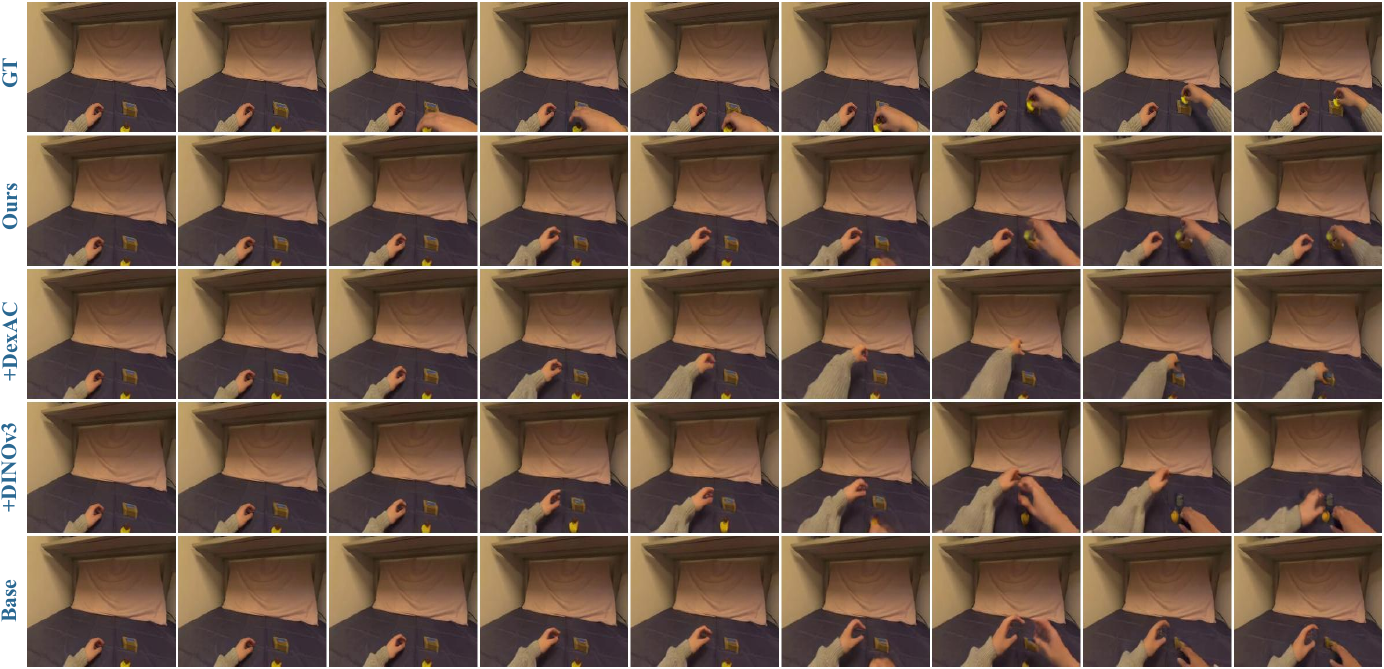}
    \vspace{-10pt}
    \caption{Stack.} \label{fig:26_8}
\end{figure*}

\begin{figure*}[t]
    \centering\includegraphics[width=\textwidth]{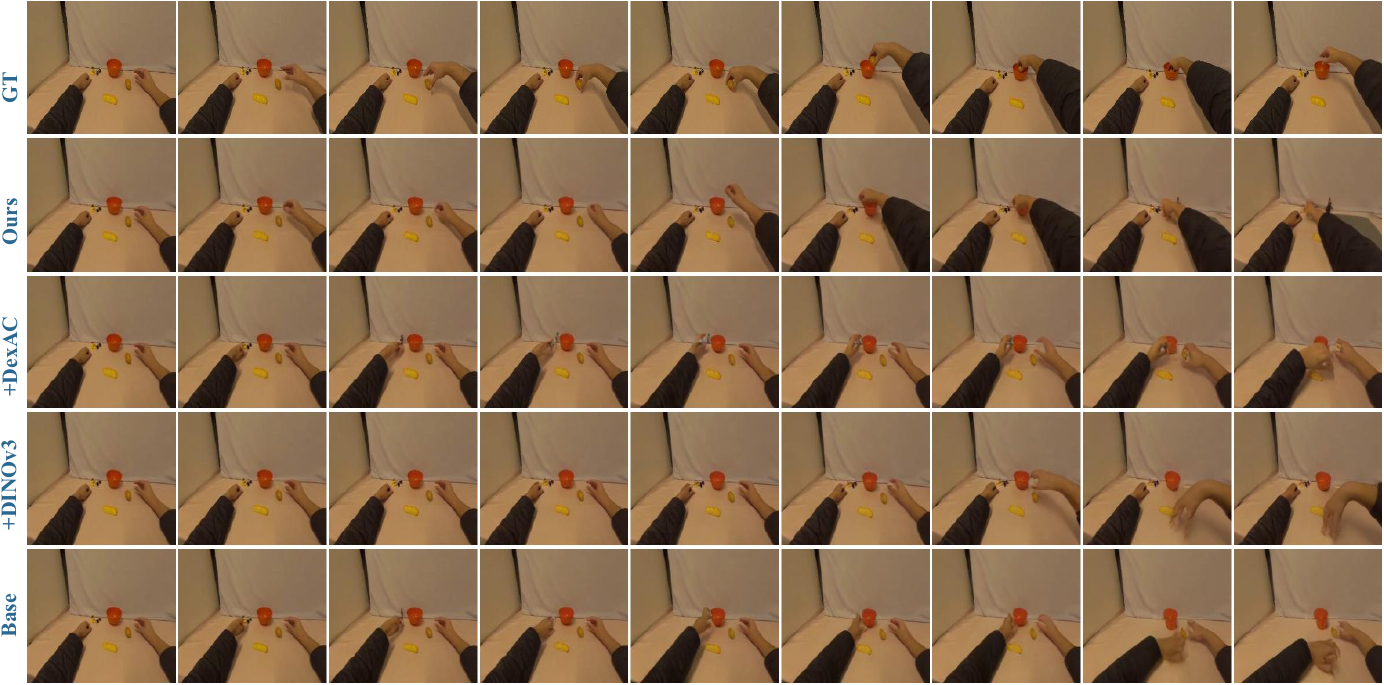}
    \vspace{-10pt}
    \caption{Basic Pick Place.} \label{fig:65_1}
\end{figure*}

\begin{figure*}[t]
    \centering\includegraphics[width=\textwidth]{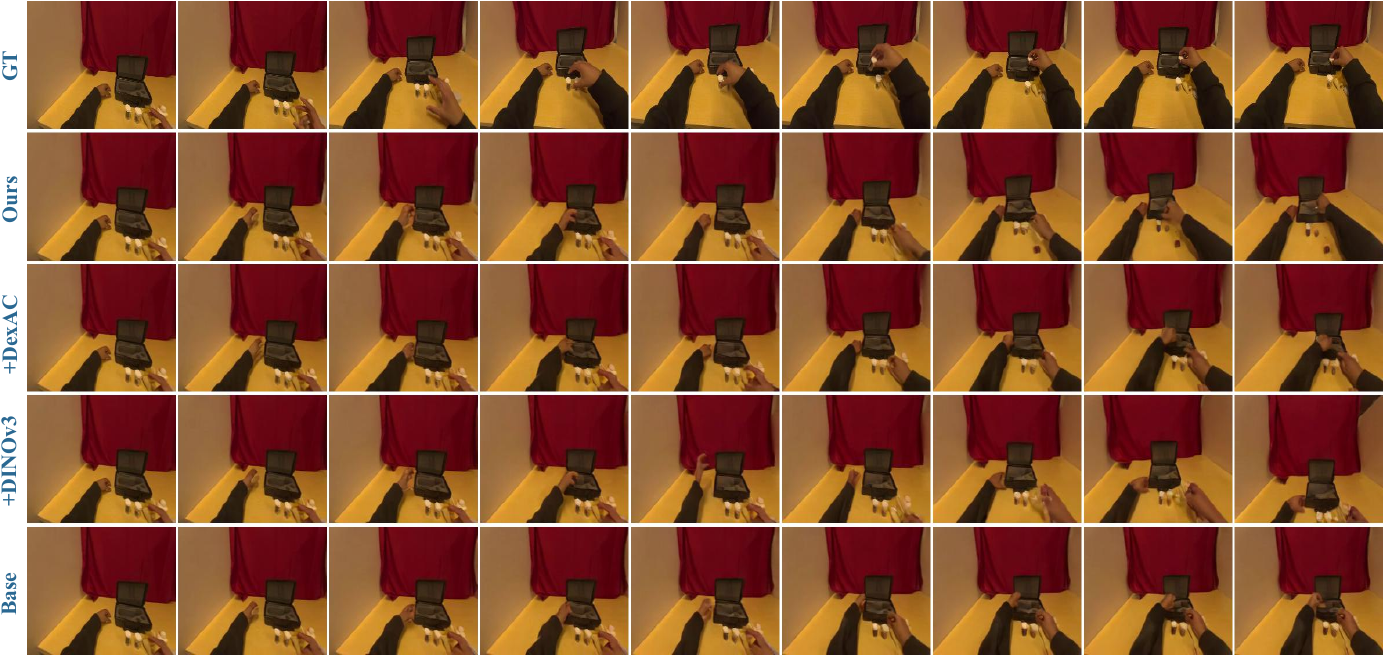}
    \vspace{-10pt}
    \caption{Open Close Insert Remove Case.} \label{fig:65_2}
\end{figure*}

\begin{figure*}[t]
    \centering\includegraphics[width=\textwidth]{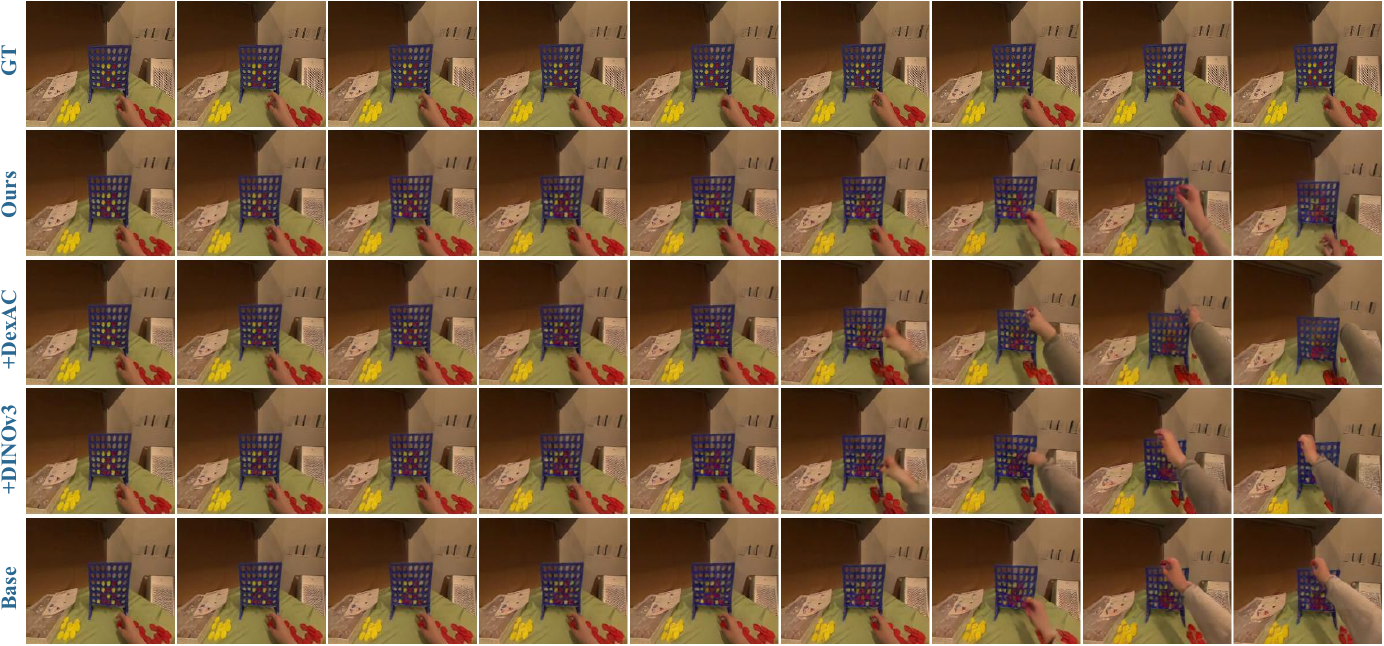}
    \vspace{-10pt}
    \caption{Play Reset Connect Four.} \label{fig:65_3}
\end{figure*}

We present additional success cases for 26-frame prediction in Figures~\ref{fig:26_1}, \ref{fig:26_2}, \ref{fig:26_3}, \ref{fig:26_4}, \ref{fig:26_6}, \ref{fig:26_7}, and \ref{fig:26_8}, and for 65-frame prediction in Figures~\ref{fig:65_1}, \ref{fig:65_2}, and \ref{fig:65_3}. To comprehensively visualize the temporal progression and long-horizon consistency of the generated interactions, we uniformly sample 8 frames at equal intervals from both the Ground Truth and predicted videos from Egodex. The Base and DINOv3 only models consistently struggle with visual-motion alignment. Although DexAC is better at capturing local hand structure representations, it exhibits macro-motion drift over long time periods. In contrast, our full method demonstrates superior spatio-temporal consistency. We observe two primary failure modes in extremely challenging scenarios shown in Figures~\ref{fig:failure_ball}. First, accumulated conditioning errors in prolonged interactions cause \textbf{long-horizon prediction deviation}. Second, self-occlusion or intricate manipulation occasionally yields \textbf{distorted finger details} due to unobserved geometry. Resolving these occlusion ambiguities and mitigating long-horizon drifts remain future directions.

\section{Limitations}
\label{sec:limitation}
The current framework mainly focuses on egocentric observations. While egocentric views provide strong embodiment signals, they restrict the model’s ability to leverage multi-view spatial information, which is important for transferring learned representations to humanoid robots and other embodied agents operating in more complex environments. Furthermore, the present study is constrained by available computational resources and thus is limited to a fixed model scale and relatively short video prediction horizons. Although this setup enables a controlled evaluation of structured action conditioning, we do not yet investigate the behavior of the proposed design under larger model capacities, longer-horizon generation, and other modal priors like VGGT \cite{wang2025vggt}. Exploring these scaling dimensions constitutes an important direction for future research. Finally, our method still depends on supervised finetuning, which is computationally expensive. Future work will focus on improving training efficiency through more lightweight adaptation, with the goal of building more scalable and spatially aware world models at lower cost.
%\section*{Appendix References} 
%\addcontentsline{toc}{section}{Appendix References}

% \begin{figure*}[t]
%     \centering\includegraphics[width=\textwidth]{meterial/26_pick_up_and_put_down_case_or_bag_ablation_comparison.png}
%     \vspace{-10pt}
%     \caption{\textbf{Pic Up and Put Down Case or Bag}} \label{fig:26_5}
% \end{figure*}

% \begin{figure*}[t]
%     \centering\includegraphics[width=\textwidth]{meterial/failure_finger_wrap_unwrap_food_9frames_comparison.png}
%     \vspace{-10pt}
%     \caption{\textbf{Failure:Finger Wrap Unwrap Food}} \label{fig:failure_finger}
% \end{figure*}
\end{document}